\documentclass[Royal,times,sagev]{sagej}

\usepackage[utf8]{inputenc}
\usepackage{graphicx}
\usepackage{amsmath,amssymb}
\usepackage{amsthm}
\usepackage{amsfonts}
\usepackage{booktabs}
\usepackage{siunitx}
\usepackage{array}
\usepackage{hyperref}
\hypersetup{hidelinks}
\usepackage{algorithm2e}
\usepackage{longtable,tabularx}
\usepackage{balance}
\usepackage{multirow}
\usepackage{subfig}
\usepackage{enumitem}

\newcommand{\tabitem}{~~\llap{\textbullet}~~}

\begin{document}

\runninghead{Fischer \emph{et al.}}
\title{Digital Simulations to Enhance Military Medical Evacuation Decision-Making}

\author{Jeremy Fischer\affilnum{1}, Mahdi Al-Husseini\affilnum{2,3}, Ram Krishnamoorthy\affilnum{1}, Vishal Kumar\affilnum{1}, and Mykel J. Kochenderfer\affilnum{2}}

\affiliation{\affilnum{1}University of California Berkeley, Fung Institute \\
\affilnum{2}Stanford University, Stanford Intelligent Systems Laboratory \\
\affilnum{3}United States Army, Department of Aviation Medicine}

\corrauth{Mahdi Al-Husseini, Stanford University, 496 Lomita Mall, Stanford, CA 94305, USA}
\email{mah9@stanford.edu}

\begin{abstract}
\textbf{Background}
The United States military medical evacuation mission is responsible for expediently evacuating the battlefield ill and injured. Medical evacuation planning involves constructing a robust network of medical platforms and facilities capable of moving and treating large numbers of casualties. We introduce the first known medium to simulate these networks in an educational setting and evaluate both offline planning and online decision-making performance. The Medical Evacuation Wargaming Initiative (MEWI) is a custom-built, high-fidelity multiplayer simulation that models tactical-level medical evacuation operations. 

\textbf{Method}
This study demonstrates the impact of simulation-based training on medical evacuation decision-making. We visualize performance data collected from two MEWI iterations executed in the United States Army’s Medical Evacuation Doctrine Course. We consider post-simulation Likert survey data from participants and external observer notes to identify key planning decision points, document medical evacuation lessons learned, and quantify general utility.

\textbf{Results}
The results highlight that participation in simulation-based medical evacuation scenarios substantially improves uptake of medical evacuation lessons learned and enhances co-operative decision-making.

\textbf{Discussion and Conclusion}
MEWI is a substantial step forward in the field of high-fidelity training tools for military medical evacuation education.  Our study findings offer critical insights into improving medical evacuation across the joint force. 

\end{abstract}


\maketitle

\section{Introduction}
This article describes and evaluates the design and implementation of a first-of-its-kind medical evacuation simulation for the United States Army's Medical Evacuation Doctrine Course at Fort Rucker, Alabama. Medical Evacuation Doctrine participants first learn the doctrinal fundamentals of tactical medical evacuation, then apply that knowledge in a series of progressive practical exercises. War theorist Carl Von Clausewitz frequently highlighted the strong collaborative relationship between war and games,\cite{sabin2012simulating} and military training has long relied on simulation-based learning to enhance decision-making and operational readiness. While the use of simulations in training dates back centuries, in the 1970s and 1980s, the United States Defense Science Board recognized simulations as a `revolution in training' with the potential to drastically and rapidly improve operational proficiency. DARPA's DARWARS program demonstrated how simulations and games could provide realistic decision-making experiences for military personnel.\cite{chatham2007games} \textit{The Art of Wargaming} emphasizes the value of decision-making in professional military simulations, and suggests that effective play requires participants to understand real-world problems and apply strategic thinking.\cite{perla1990art} 

Modern combat preparedness increasingly relies on digital simulations and simulation software to enhance operational decision-making and crisis response. BRAC University in Bangladesh developed a game-engine based military simulator that enabled multiplayer platform control in movement and maneuver operations.\cite{azmi2016multiplayer} A Lebanese Port Explosion simulator effectively captured the uncertainty inherent in many casualty generation scenarios.\cite{hajj2024simulating} The Combat Simulation Lab at West Point actively employs IWARS, JCATS, OneSAF, VBS2, and other advanced simulations to teach cadets doctrinal techniques related to small-scale tactical engagements.\cite{WestPointSimulationCenterUnitedStatesMil} While these simulations adequately replicate battlefield friction and uncertainty, they do not contain a meaningful medical evacuation component. 

Civilian emergency medical response training also continues to advance rapidly. Simulation-based tools involving mannequins and complex software virtual environments are increasingly common.\cite{lane2001simulation} Digital emergency care simulations have helped evaluate first-aid training while capturing the stress and chaos of an emergency scenario.\cite{rash2024virtual} Niles et al. introduce a simulation learning exercise where participants dispatch and control a set of unmanned aircraft systems responding to opioid overdoses in rural areas.\cite{niles2025emergency} Closely related to military medical evacuation is dispatch training for civilian emergency medical services. Yuan and Lam discuss optimizing ambulance dispatch on a by-casualty basis and how to quantify and score response times. The authors reveal a seemingly unintuitive inverse relationship between patient transport times and ensuring continuous evacuation system support.\cite{yuan2018simulator} Similarly, balancing evacuation expediency with evacuation coverage is a fundamental tradeoff that regularly challenges participants in the Medical Evacuation Doctrine Course. Our proposed simulation, the Medical Evacuation Wargaming Initiative (MEWI), combines sound military wargaming principles with recent advancements in civilian emergency medical education to both reinforce doctrinal medical evacuation knowledge and support critical thinking in complex and evolving evacuation scenarios.

Within the specific domain of military medical evacuation, a substantial body of operations research has established rigorous methods for optimizing asset management. Foundational work by Bastian,\cite{bastian2010robust} Grannan,\cite{grannan2015maximum} and Fulton \cite{fulton2010two} focuses largely on the strategic location and allocation of assets, utilizing stochastic programming and Monte Carlo simulations to determine optimal force structures for steady-state stability operations in Afghanistan and large-scale combat scenarios. Complementing this work, Jenkins,\cite{jenkins2021approximate, jenkins2018examining} Rodriguez,\cite{rodriguez2023solving} and Frial \cite{frial2024characterizing} have extensively examined the operational challenges of dispatching, routing, and re-routing platforms. Their research frequently employs Markov Decision Processes and Approximate Dynamic Programming to overcome the `curse of dimensionality' inherent in dynamic decision-making, testing their algorithms against complex problem sets to include high-intensity conflict scenarios in Azerbaijan and Korea. Together, these works provide a powerful analytical framework for determining high-quality policies for the placement and movement of medical evacuation platforms under uncertainty.

However, a critical distinction exists between these analytical optimization models and the simulation proposed here. The primary objective of the aforementioned literature is developing solvers that can computationally derive efficient policies within tractable mathematical bounds. In contrast, MEWI focuses on human-centric approaches, where the `solver' is the participant. Because MEWI is not constrained by algorithmic tractability, it can support a world model of significantly higher fidelity, incorporating complex terrain, fog of war, and emergent multi-agent friction that mathematical models must often abstract away. While previous work answers the question of how an optimal system should behave, MEWI addresses the pedagogical challenge of training and evaluating human policies in the chaos of a simulated environment.

\section{Medical Simulation Design}

\begin{figure}[t]
\centering
\includegraphics[width=\linewidth]{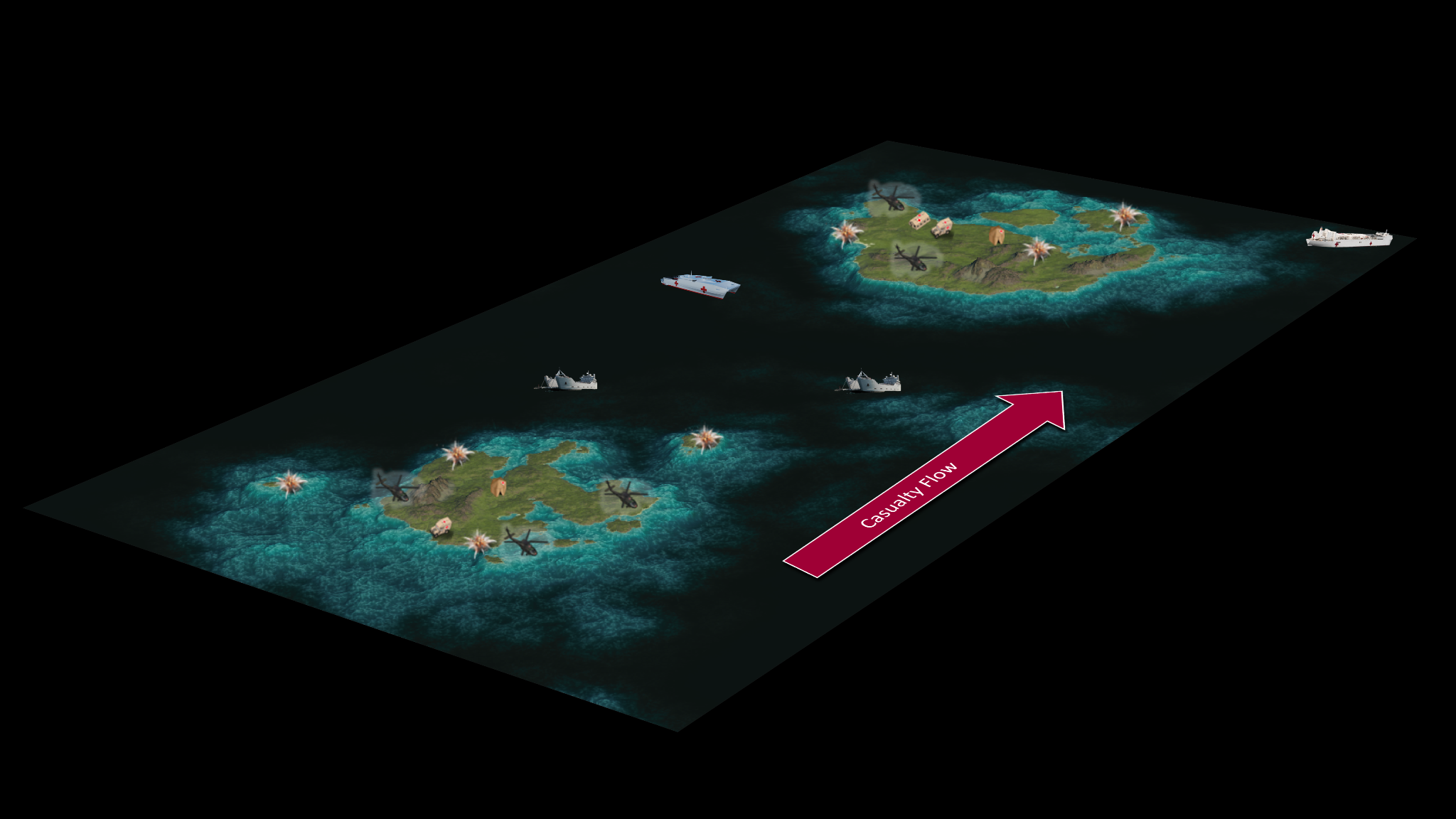}
\caption{A representative island-based medical evacuation scenario.}
\label{fig:maritimescenario}
\end{figure}

\subsubsection{Simulation Objectives} MEWI was developed for the United States Army's Medical Evacuation Doctrine Course at the Department of Aviation Medicine in partnership with the University of California Berkeley Fung Institute. MEWI seeks to enhance educational outcomes by providing insight into conducted medical evacuation planning, reinforcing medical evacuation doctrine-based learning, and fostering critical thinking in groups. Medical evacuation planning for complex operational environments, as shown in Figure \ref{fig:maritimescenario}, is challenged by limited and diverse evacuation platforms, substantial and evolving casualty transport requirements, enemy threat considerations, partial system observability, expansive terrain, joint service integration, and other constraints. MEWI is the first high-fidelity three-dimensional digital simulation tool dedicated to improving military medical evacuation decision-making in an educational setting. Examples of evacuation decisions made in MEWI include: how to move patients from various collection points to increasingly sophisticated medical treatment facilities, which platforms will facilitate various legs of this movement, how to triage patients throughout the evacuation process, how to balance limited bed space across the battlefield and minimize transport delays, and how to document patient information and status in space and time.\cite{leek2025case}

\begin{figure}[t]
\centering
\includegraphics[width=0.65\linewidth]{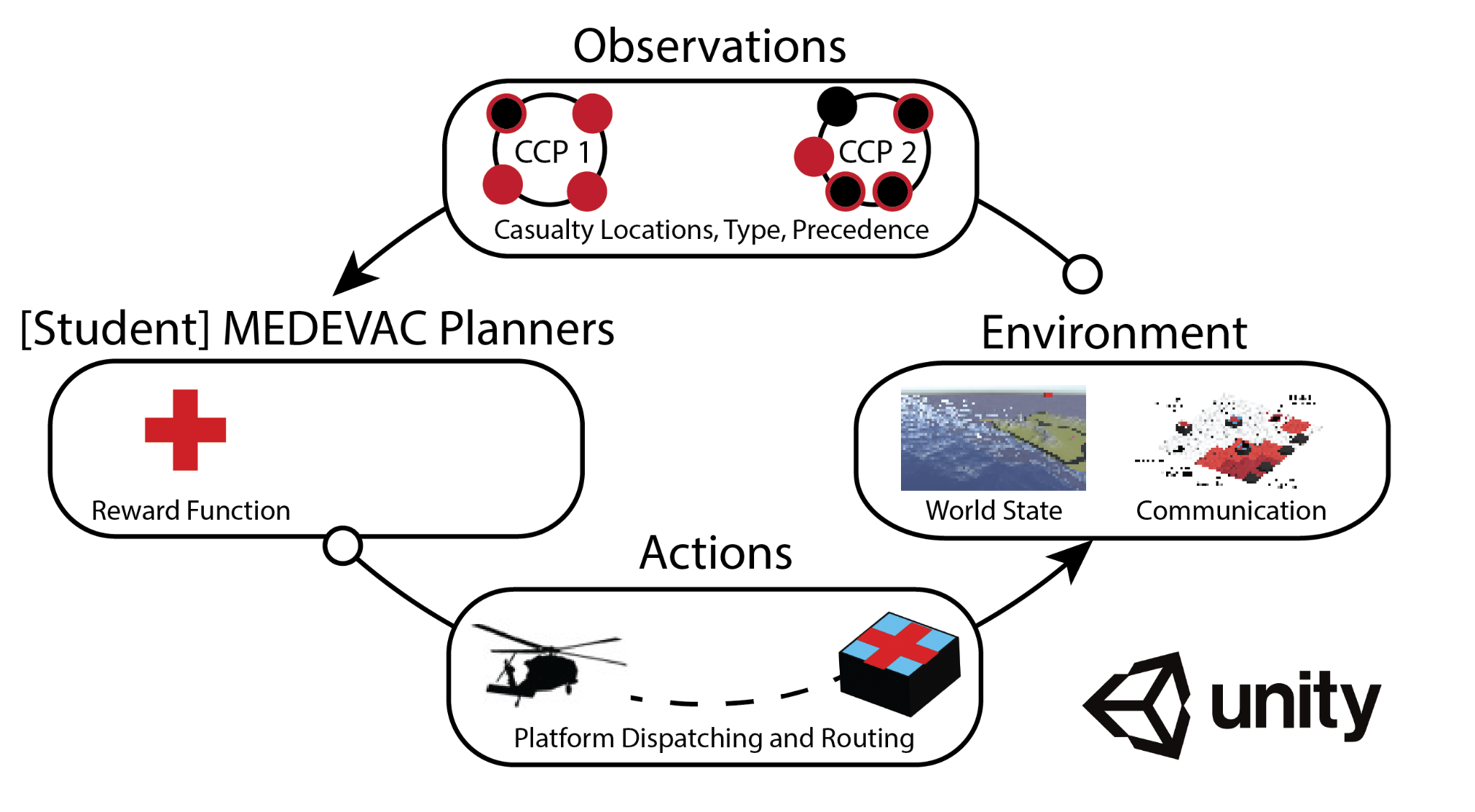}
\caption{The Markov Decision Process inspired model for simulation execution.}
\label{fig:MDP}
\end{figure}

\subsubsection{Model Design} We model the interactive MEWI environment, shown in Figure \ref{fig:MDP}, using a multi-agent, sometimes partially observable, and often semi-Markovian variant of the Markov Decision Process (MDP).\cite{kochenderfer2022algorithms, bellman1957markovian} The MDP is a foundational fully cooperative stochastic decision-making model regularly used in the artificial intelligence sub-fields of planning and learning. At a minimum, this model includes states, a state transition function, agents, their actions, and a reward function. We discuss specific elements of the model below.

\subsubsection{Agents} Each player is an agent with centralized control over one or more evacuation platforms. We therefore use the terms player and agent interchangeably. Evacuation platforms may be ground vehicles, helicopters, or ships. Each platform has well-defined constraints and capabilities and is dispatched and routed in the shared environment to support a collaborative objective. The actions of one agent affect the actions of another. For example, if two players attempt to land a helicopter on the same single-pad ship, the second to arrive will enter a waiting queue. This is a form of model transition dependence.

\subsubsection{Rewards} Players are rewarded for minimizing evacuation transport times. Each patient maintains a vector $v = [T_0,T_1,T_2,T_3]$, where $T_0$ is the initialization time, and subsequent components represent arrival times at military medical facilities also known as the roles of care. A Role 1 typically contains two to three medical professionals with limited medical equipment and is primarily concerned with initial trauma care and forward resuscitation, whereas a Role 3 may be a large field hospital with surgery and can accommodate hundreds of patients. A Role 2 has manning and capabilities somewhere between a Role 1 and Role 3. $T_1$ is the arrival time at the Role 1, $T_2$ is the arrival time at the Role 2, and $T_3$ is the arrival time at the Role 3.

\begin{equation}
p_{\text{max}}\left(l - \frac{E_s}{T_2 - T_0}\right)
\label{eq:eq1}
\end{equation}

Patients are initialized with a maximum $p_{\text{max}}$ score, which decreases linearly with evacuation time to the Role 2. In Eq \ref{eq:eq1}, $l$ is initialized to ten for urgent patients and eight for priority patients to prioritize those requiring immediate evacuation assistance and treatment. $E_s$ is the NATO doctrinal evacuation standard time of one hour for urgent patients and four hours for priority patients. Two mortality curves with exponential distributions are sampled each time an urgent patient is instantiated, resulting in two unique death times $T_{\text{Death, 1}}$ and $T_{\text{Death, 2}}$. A patient dies if $T_1 - T_0 > T_{\text{Death, 1}}$ or if $T_2 - T_1 > T_{\text{Death, 2}}$. Role 1 care provides immediate patient stabilization but lacks the medical resources to guarantee patient survival. This motivates maintaining $T_{\text{Death, 2}}$, which encourages players to rapidly facilitate further transport to the Role 2. The mortality curve parameters align with military medical literature on the physiological `Golden Hour' principle and historical outcomes in combat environments.\cite{kotwal2016effect, parker2007casualty, newgard2010emergency} Each death results in a -10 penalty and supplants $p_{\text{max}}$. Points accumulated across casualties are revealed at the conclusion of the simulation on the score screen. Priority patients are not assigned a mortality curve in simulation. Consistent with NATO doctrine, priority patients require prompt transfer for specialized treatment not available locally, but injuries are not immediately life-threatening. Evacuation is needed within a reasonable time frame to prevent unnecessary pain or disability.

\subsubsection{Actions}: Agent actions reflect realistic medical evacuation decision-making and include:
\begin{itemize}
    \item Aligning evacuation platforms to evacuation requests.
    \item The routing of evacuation platforms in the environment.
    \item Picking up or dropping off casualties at Role 1, 2 or 3s.
    \item Organizing patient transfers at exchange locations to include Ambulance Exchange Points (AXPs) and Helicopter Landing Zones (HLZs).
    \item Waiting, as patients cannot be left alone at exchange locations, and transfer times may require platform loitering.
\end{itemize}

\subsubsection{States}: Each state maintains information relevant to player decision-making at each time step. The initial state is relayed to participants prior to simulation execution through scenario-specific operation orders, briefings, and previous class instruction. Most state information is however observed online during game-play. Agents know their platform locations and capacities, as well as key casualty information including patient precedence (urgent or priority), patient type (litter or ambulatory), and current location. Other known elements include the location of all exchange points, other agent platforms, and threat zones. State information is readily accessible to participants through the Unity game engine interactive multiplayer interface. It is critical to player situational awareness and informed decision making that relevant state variables are incorporated into an intuitive and accessible user interface. 

\subsubsection{Transitions}: State transitions are determined using a mix of stochastic and deterministic processes conditioned on agent and adversarial actions. The adversarial agent places ground obstacles and air defense rings over transit routes and destination points, causing arrival times and planned routes to shift. Casualty generation at collection points follows a stochastic process, introducing randomness to when and how many patients are produced. While recent literature suggests Hawkes processes better model the self-exciting nature of firefights \cite{frial2024characterizing}, we instead use a Poisson distribution to model the aggregate arrival of casualties across a large operating environment. This provides a consistent, stochastic demand signal sufficient to stress student decision-making and validate adherence to doctrine without the added computational complexity of self-exciting models. We instead choose to induce mass casualty events via on-demand facilitator injects.
\subsubsection{Observations}: State information may be intentionally hidden from players, whether due to fog of war constraints or the nighttime cycle. This forces players to plan using incomplete or noisy observations of the underlying state, which can result in suboptimal decision-making. Partial observability encourages players to communicate their observations with one another outside of the confines of the simulation. Agent coordination is critical to success, as communication heavily influences the extent to which the underlying state is observed and can be acted upon.

\section{Scenarios}

\begin{figure}[t]
\centering
\includegraphics[width=0.99\linewidth]{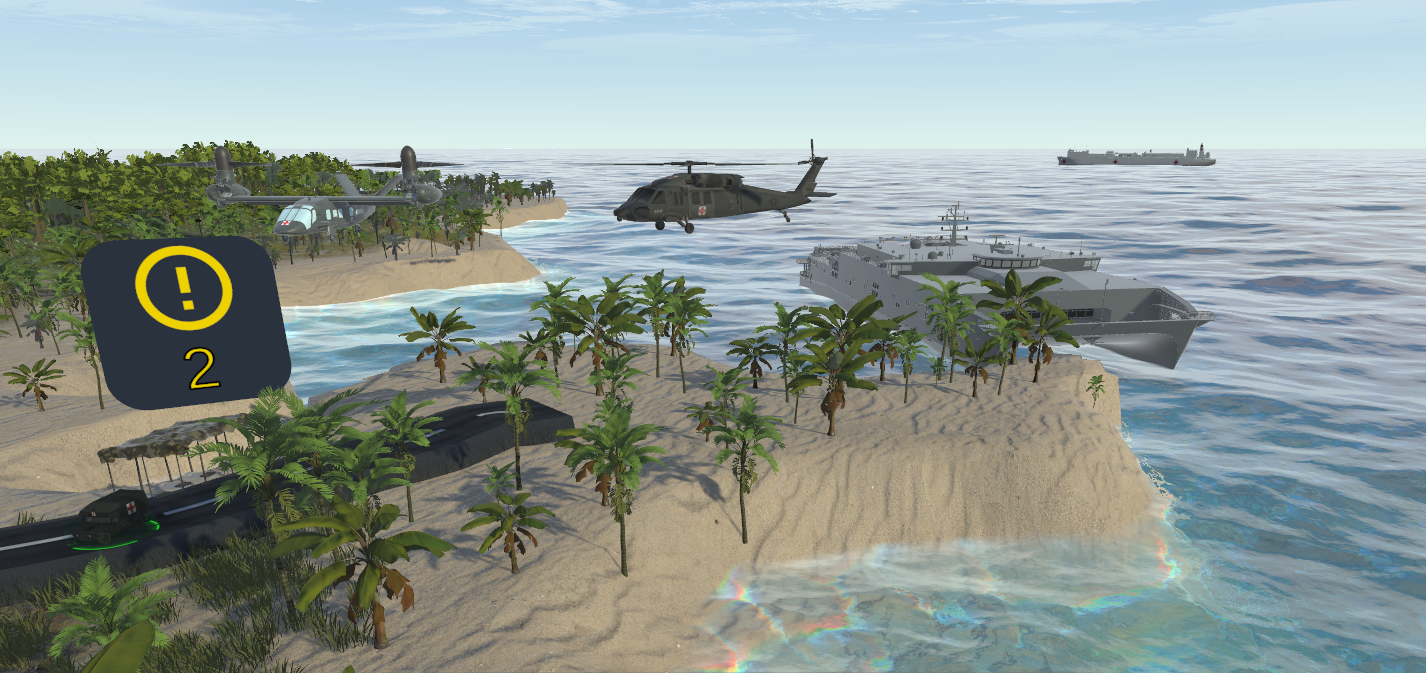}
\caption{Various multi-modal evacuation platforms dispatched and routed by participants.}
\label{fig:platforms}
\end{figure}

As stated in Force Design 2030, \cite{harper2021marine} the United States Marine Corps is actively reverting its force structure to its roots as an amphibious assault organization. Similarly, the United States Army is adjusting training, manning, and material to support littoral operations. The first simulation scenario, Operation Storm Surge, was therefore designed to reflect a joint amphibious assault taking place on the Hawaiian Islands. Operation Storm Surge generates casualties at seven casualty collection points (CCPs) selected  
across two islands. Agents must coordinate multi-modal medical evacuation platforms, shown in Figure \ref{fig:platforms}, to facilitate patient evacuation from the islands to offshore hospital ships. Operation Storm Surge is characterized by mass casualty events, which results in a substantial number of casualties being generated at select CCPs in a short period of time.

\begin{figure}[t]
    \centering
    \subfloat[\centering ]{{\includegraphics[width=0.49 \linewidth]{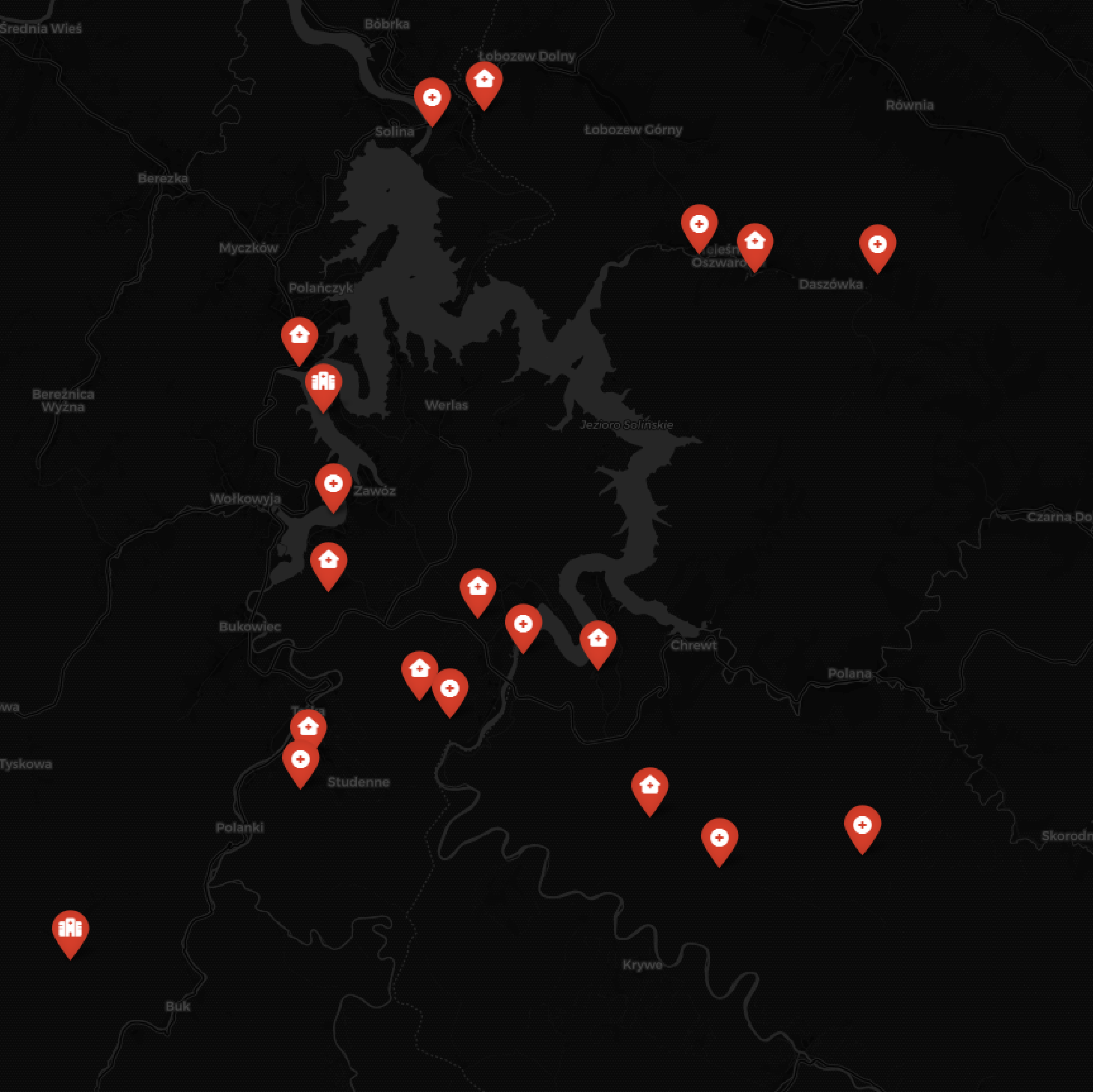} }}%
    \subfloat[\centering ]{{\includegraphics[width=0.49 \linewidth]{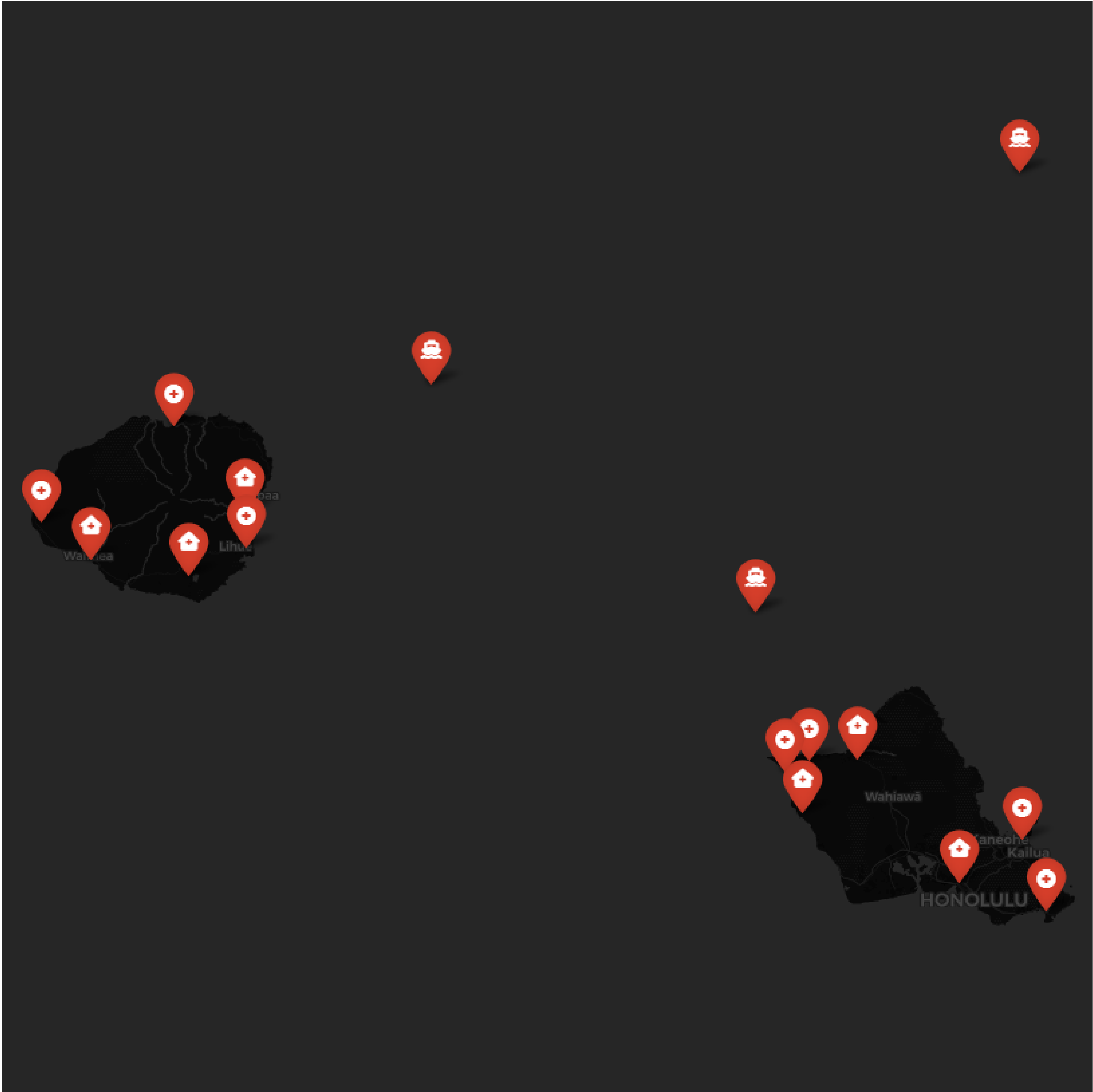} }}%
    \caption{Operation Eastern Crucible (a) and Storm Surge (b) scenario layouts of CCP (cross in circle), Role 1 (cross in house), Role 2 (ship/hospital), and Role 3 (ship/hospital) locations. Participants determine AXP, HLZ, Role 1, and Role 2 locations during the planning phase.}
    \label{fig:scenarios}
\end{figure}

The second simulation scenario, Operation Eastern Crucible, takes place in Eastern Europe and is inspired by the Russia-Ukraine conflict. The environment is defined by vast terrain, several waterways, and an extensive road network critical for maneuver and sustainment. Unlike the island-hopping requirements in the Pacific scenario, this conflict is primarily land-based, with combined arms forces navigating contested highways, rural landscapes, and a shifting front line. CCPs activate and deactivate as the battle progresses, forcing participants to adapt their medical evacuation and treatment strategies, and extend their evacuation network, in real-time. Participants must effectively integrate ground and air medical evacuation platforms while planning against changing CCP locations to ensure timely and effective care in a rapidly evolving operational environment. Features for both scenarios are documented in Table 1 and collection point and treatment facility locations are shown in Figure \ref{fig:scenarios}.

\begin{table}[h]
\scriptsize
\centering
{
\begin{tabular}{|p{3cm}|p{4cm}|p{4cm}|}
\hline
\textit{Category} & \textbf{Operation Storm Surge} & \textbf{Operation Eastern Crucible} \\ \hline
Geography & Maritime amphibious assault on the Hawaiian Islands of Oahu and Kauai, with dense terrain. & Combined arms offensive across expansive road network with wet-gap crossings and contoured terrain.\\ \hline
Operation Duration & 24 hour operation across three phases, beginning at 1600 LCL. & 24 hour operation across three phases, beginning at 1100 LCL.
\\ \hline
Platforms/ Medical Facilities & 
\tabitem 1x Role 3 US Mercy hospital ship 

\tabitem 2x Role 2 EMS (routable)

\tabitem 2x Bell MV-75s

\tabitem 3x HH-60M Black Hawks

\tabitem 16x M997A2 Ground Ambulances
&
\tabitem 1x Role 3 Field Hospital

\tabitem 1x Role 2 BSMC

\tabitem 6x HH-60M Black Hawks

\tabitem 20x M997A2 Ground Ambulances

\tabitem 4x M1133 MEVs

\tabitem 8x  M113A4s
\\ \hline
CCP Characteristics & All seven CCPs active simultaneously by mid-scenario. & Nine CCPs dynamically activate and deactivate at the start of each operational phase. 
\\ \hline
Threat Conditions & \multicolumn{2}{|c|}{Adjustable enemy air defense rings, ground obstacles, and artillery.} 
\\ \hline
Learning Focus & 
Coordinating joint evacuation operations using maritime, air, and ground platforms across non-contiguous terrain. Execution of Mass Casualty Plan. &
Employing ground-based shuttle systems with effective air-ground integration while contending with a moving Forward Line of Own Troops (FLOT). 
\\ \hline
\end{tabular}
}
\caption{Scenario comparison}
\label{tab:comparison_scenarios}
\end{table}

\section{Features} We introduce three categories of features: limited information, policy constraints, and facilitator injects, to emulate a realistic environment for medical planners and evacuation platform operators.

\subsubsection{Limited Information:} 
\begin{figure*}[t!]
    \subfloat[\centering Operations during Dusk]{%
        \includegraphics[width=.49\linewidth]{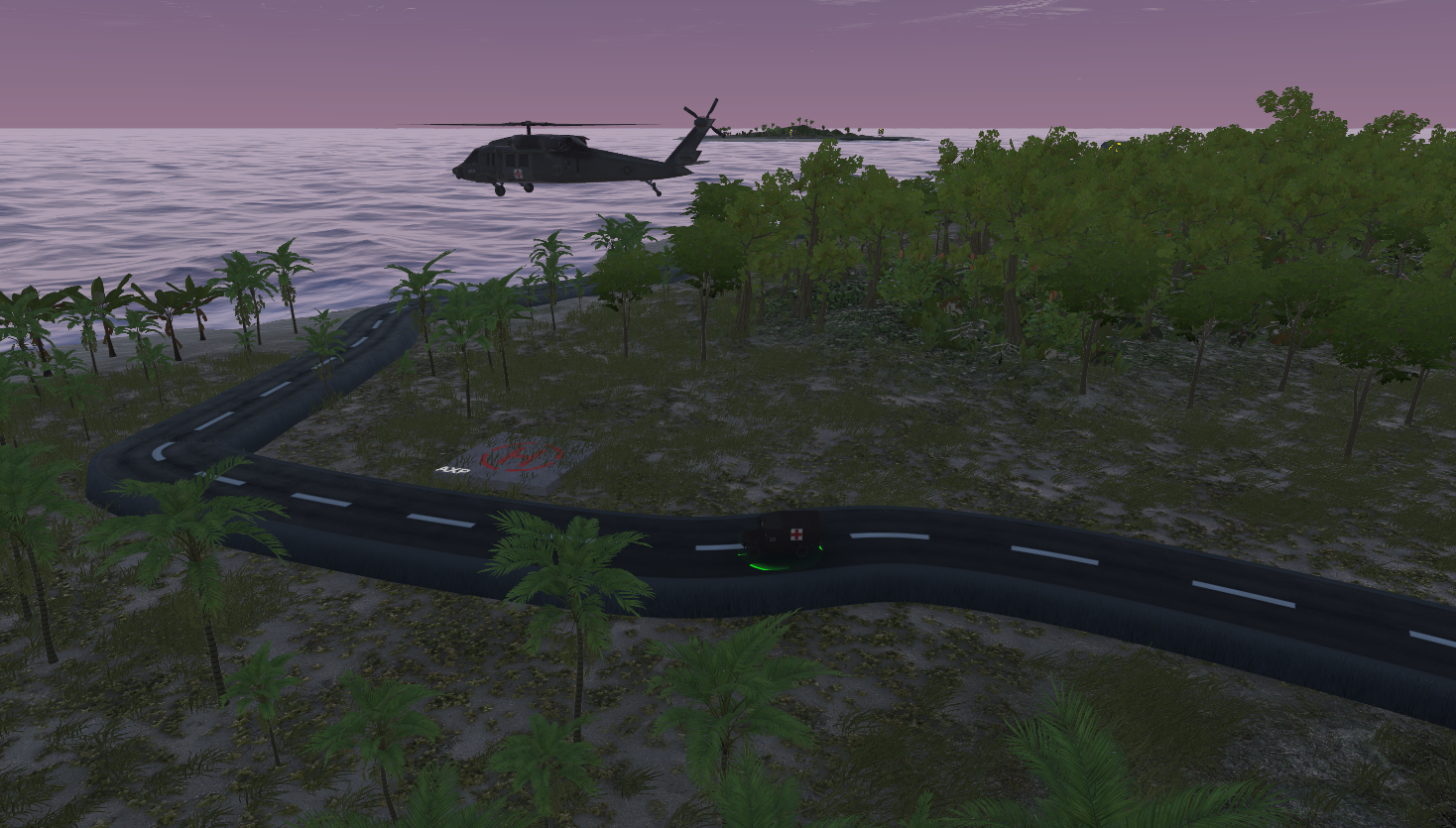}%
        \label{subfig:a}%
    }\hfill
    \subfloat[\centering Enemy Artillery]{%
        \includegraphics[width=.49\linewidth]{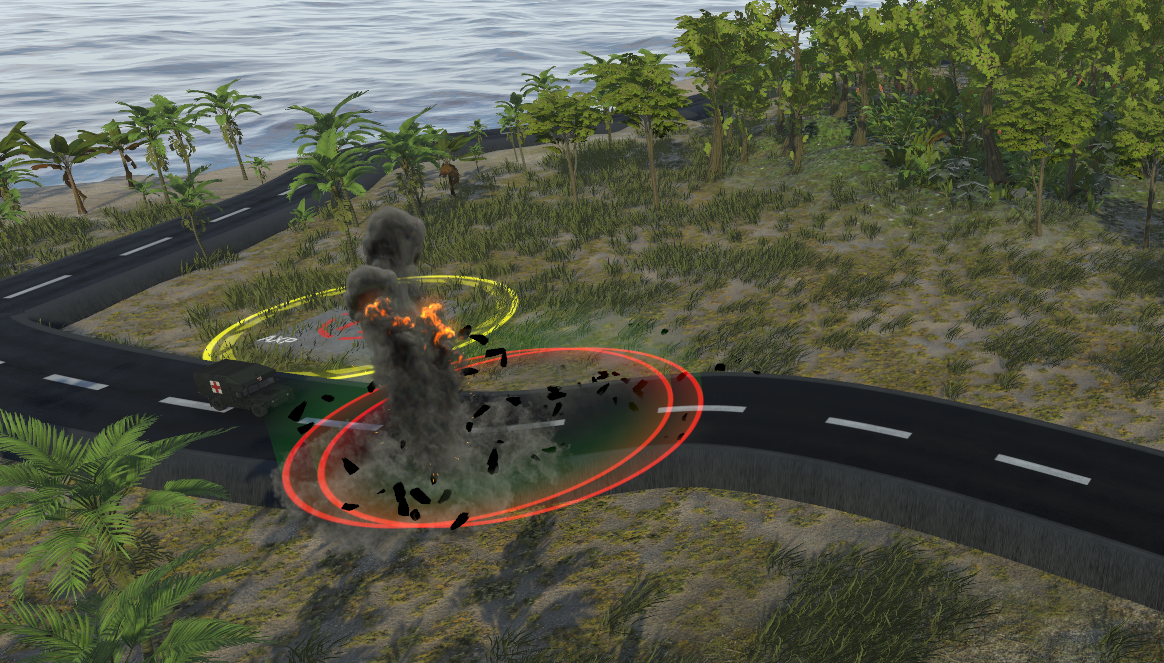}%
        \label{subfig:b}%
    }\\
    \subfloat[\centering Ground Obstacles]{%
        \includegraphics[width=.49\linewidth]{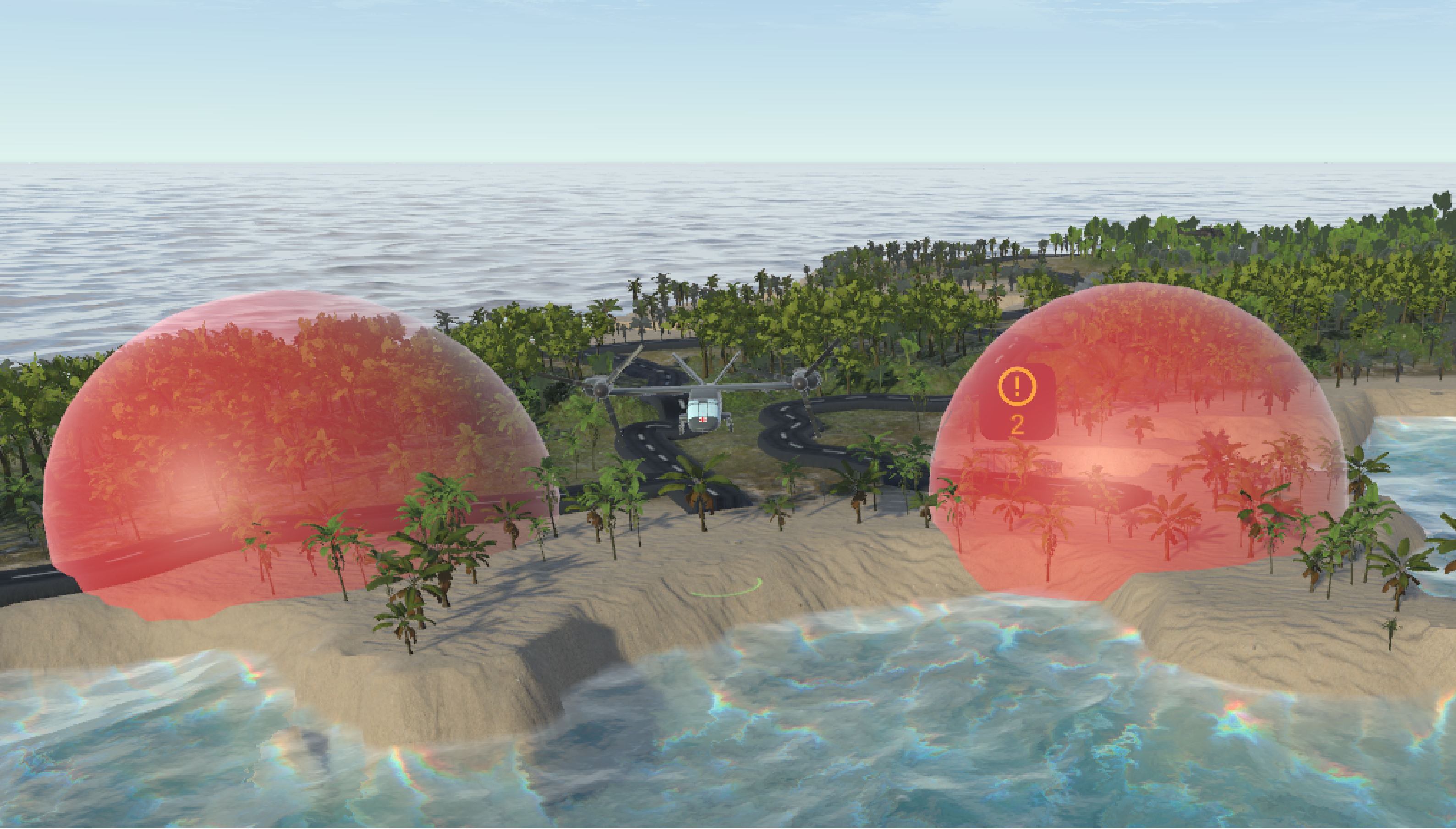}%
        \label{subfig:c}%
    }\hfill
    \subfloat[\centering Air Defense Rings]{%
        \includegraphics[width=.49\linewidth]{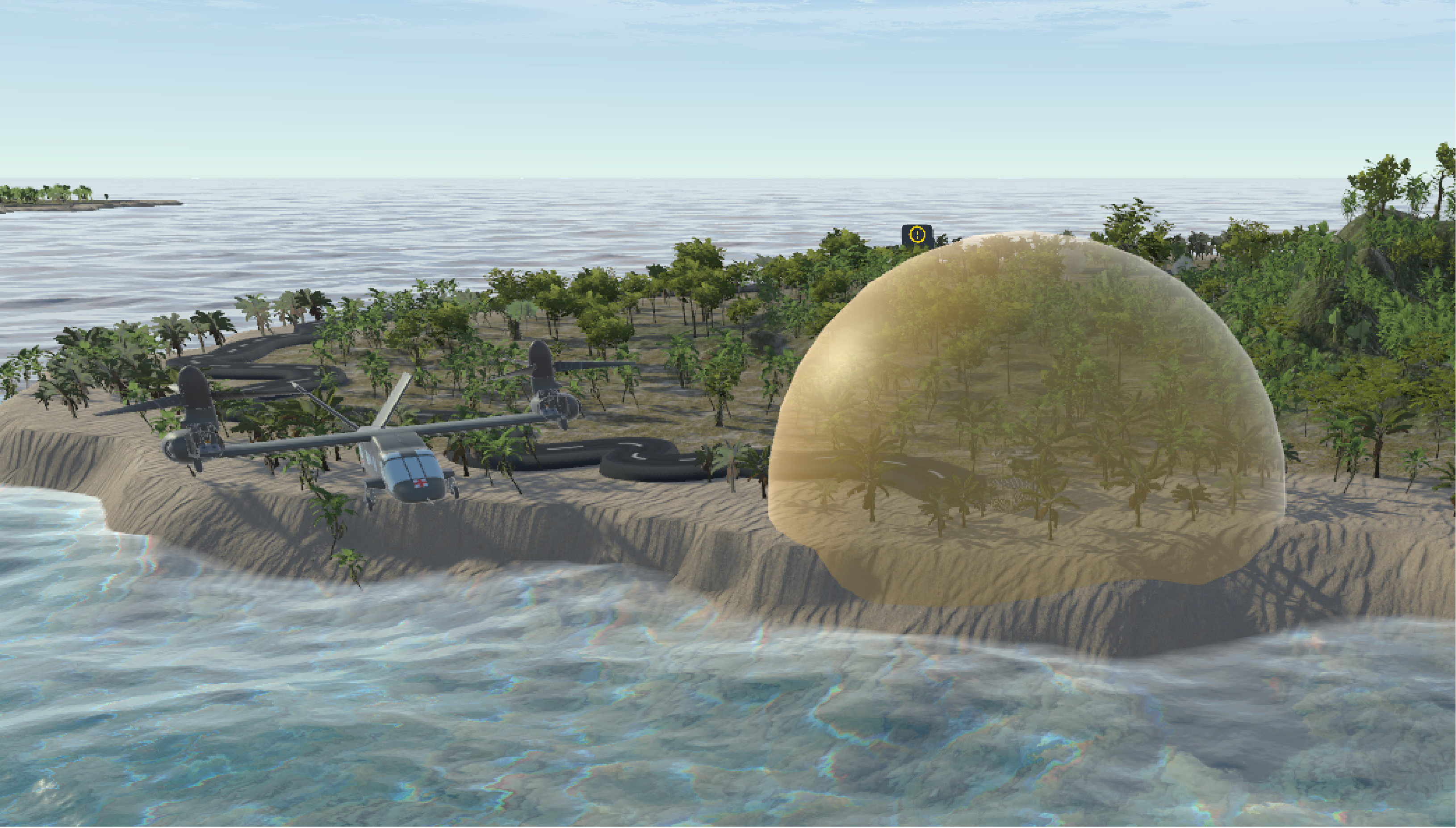}%
        \label{subfig:d}%
    }
    \caption{Various features available as part of scenario Operation Storm Surge.}
    \label{fig:features}
\end{figure*}

Decision-making under uncertainty is challenging. Warfighting typically involves some measure of `fog of war', or state uncertainty. Medical planners scarcely receive curtailed instructions, receiving much of the same operational information as combat arms troops, often in the form of a written document called an operations order. Similarly, simulation participants enter the planning phase with only general information about the upcoming operation conveyed in a scenario-specific operations order. After planning, participants are permitted to place their platforms, exchange locations, and medical facilities on the digital map. While the environment is dynamic, fixed site locations are typically maintained in their initial position to provide predictability and enable coordination, while employment of those sites evolves in response to the casualty stream. Features that contribute to state uncertainty include: 

\begin{itemize}
    \item \textbf{Day and Night Cycle} Modeling the day to night transition, during which the simulation environment darkens, restricts visibility and challenges coordination between participants. An example environment during dusk hours is shown in Figure \ref{fig:features} (a).
    \item \textbf{Adversarial Artificial Intelligence} We use reinforcement learning algorithm Proximal Policy Optimization (PPO) \cite{yu2022surprising} to inform an adversarial agent capable of intelligently placing ground obstacles and air defense rings, as seen in Figure \ref{fig:features} (c) and (d). Both ground obstacles and air defense rings are updated with an adjustable frequency by the adversarial agent such that friendly movement on roads networks and in airspace is obstructed, therefore impeding casualty collection. The adversarial agent is also modeled as an MDP, and receives a reinforcing reward whenever it meaningfully delays patient movement to successive roles of care. Participants cannot predict these obstructions and must adjust their actions accordingly in real-time.
    \item \textbf{Stochastic Casualty Generation} Casualties are generated using Poisson distributions, with $\lambda$$_1$ representing the time between casualty incidents and $\lambda$$_2$ the number of casualties per incident. Parameters may be conditioned on a selected difficulty level and the tactical scenario. Additionally, CCPs generate casualties at different, hidden rates (e.g., CCP 1 may generate twice as many casualties as CCP 3 based on the tactical scenario), further complicating planning. 
\end{itemize}

\subsubsection{Policy Constraints:} The simulation incorporates several established medical evacuation policy constraints.

\begin{itemize}
    \item \textbf{One Hour Evacuation Standard} Scoring is governed by policy, including the one-hour evacuation standard for transporting urgent patients from CCPs to Role 2 facilities. This may be modified by simulation facilitators to help participants evaluate the impact of various evacuation policies in different scenarios.
    \item \textbf{Platform Capacity and Patient Type} All air, ground, and maritime platforms maintain their actual capacity limits, including distinctions between ambulatory (walking wounded) and litter (stretcher-bound) patients. Patient loading and unloading times also match doctrinal estimates. 
    \item \textbf{Progressive Movement} Consistent with the Army doctrine,\cite{medicalevac} patients must transit through each progressive role of care sequentially: Role 1, Role 2, and then Role 3. Facilitators may approve exceptions to progressive movement on a by-patient basis if appropriately justified by participants. These exceptions exist for certain critical patients and for transport on non-contiguous battlefields. 
    \item \textbf{Continuity of Care} Patients cannot be left unattended. Evacuation platform dropping off patients at control measures like AXPs and HLZs cannot depart until another platform has arrived to take the patient. The game enforces this by restricting platform movement until the patient transfer is complete. 
    \item \textbf{Aircraft Accommodation Limits} The Operation Storm Surge Role 2 facilities are USNS Expeditionary Medical Ships (EMS) and the Role 3 is the USNS Comfort hospital ship. Both medical and hospital ships can accommodate a single aircraft due to only having one landing pad. If multiple aircraft are present at any one ship for transport, subsequent aircraft must queue until the first to arrive has completed its transfer. 
    \item \textbf{Realistic Scale} After building the terrain for both scenarios using height maps, the environments were refined with realistic road networks, medical facilities, and movement constraints that reflected each platform’s capabilities. The accurate scaling of time, distance, and elevation ensures that participants experience a realistic sense of control and situational awareness.
    \item \textbf{Casualty Considerations} MEWI explicitly models urgent and priority casualties but excludes routine casualties. This design choice reflects the reality of Large-Scale Combat Operations (LSCO), where the evacuation system is saturated by life-threatening cases. Routine patients are assumed to be managed via return-to-duty (RTD) protocols at the unit level or up the Role 1, rather than consuming limited evacuation assets.
\end{itemize}  

\subsubsection{Facilitator Injects:} Injects permit facilitators to add realism into any evacuation scenario and stress participant decision-making.

\begin{itemize}
    \item \textbf{Enemy Artillery} As seen in Figure \ref{fig:features} (b), facilitators can range friendly medical platforms with artillery. A targeted platform is immobilized for a specified period of time and all patients in the platform are then killed. Facilitators may issue tactical prompts which then appear as banner text on all participant screens. Operating in a manner that does not align with issued prompts may be justification for a facilitator to fire artillery at participant platforms. Similarly, facilitators may restrict platforms to specified operating zones in the planning phase. For example, medical ships may be prohibited by policy from being moved within 10 miles of any beachhead. Participants ships that near the beachhead may then be subject to artillery. 
    \item \textbf{Casualty Flow Management} Facilitators can toggle CCPs on or off if casualty volume becomes unmanageable, allowing recalibration of simulation difficulty to maintain instructional value. Additionally, facilitators can generate mass casualty events at select CCPs to intentionally stress the evacuation system at perceived pain points.
    \item \textbf{Casualty Evacuation Platforms} Facilitators possess casualty evacuation platforms like the CH-47 Chinook helicopters and Light Medium Tactical Vehicle (LMTV) to assist with patient movement. These platforms must be requested by participants for specified numbers of lifts and with proper justification. This feature helps demonstrate the need to strategically integrate medical and casualty evacuation platforms. Unlike a medical evacuation platform, a casualty evacuation platform does not necessarily have en-route medical care, is not dedicated to patient movement, and is not marked with red cross insignia. 
    \item \textbf{Real-Time Scoring and Performance Monitoring} Facilitators have access to live scoring and tracking metrics to include patient deaths and transport times. This allows facilitators to assess participant performance and adjust injects accordingly.
    \item \textbf{Communication Blackout} Facilitators may induce communication blackouts, during which participants cannot verbally communicate with one another. This limits observation sharing, complicates the execution of joint tasks like patient exchanges, and generally challenges medical evacuation decision-making.
\end{itemize}

\section{Participants and Positions}

\begin{figure}[t]
\centering
\includegraphics[width=0.70\linewidth]{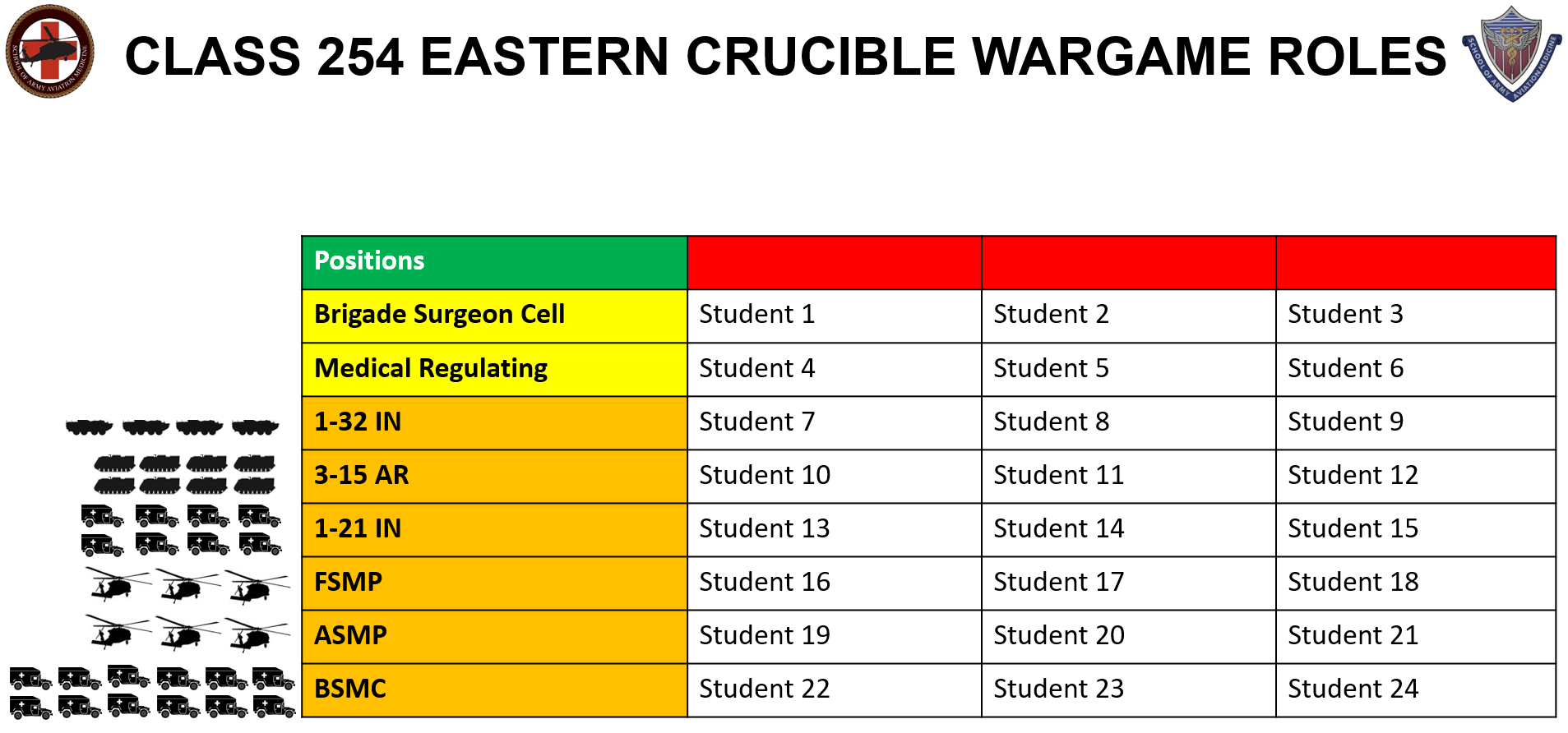}
\caption{Simulation roles for Operation Eastern Crucible with assigned platforms.}
\label{fig:positions}
\end{figure}

Figure \ref{fig:positions} shows how each participant is assigned a position associated with the tactical medical evacuation plan for a given scenario. Positions in orange possess evacuation platforms that need to be allocated and dispatched in the simulation, while positions in yellow are concerned with command and control and information management. Each position is discussed during the simulation tutorial and during the road to war briefing. The types of medical evacuation decisions made in planning differ by position, and decisions made affect the collective. 1-32 IN, 3-15 AR, and 1-21 IN are examples of ground maneuver battalions, the forward support medical evacuation platoon (FSMP) and area support medical evacuation platoon (ASMP) are responsible for aeromedical evacuation support, and the brigade support medical company (BSMC) is effectively the Role 2 medical treatment facility. Medical regulators must collect location and medical condition information on all patients transiting the evacuation system in real-time. The brigade surgeon is the overall medical authority for the simulation and is responsible for ensuring the evacuation system runs smoothly. In larger groups, participants may be separated into two distinct teams competing against each other.

\section{Instructional Setting}

\begin{figure}[t]
\centering
\includegraphics[width=0.70\linewidth]{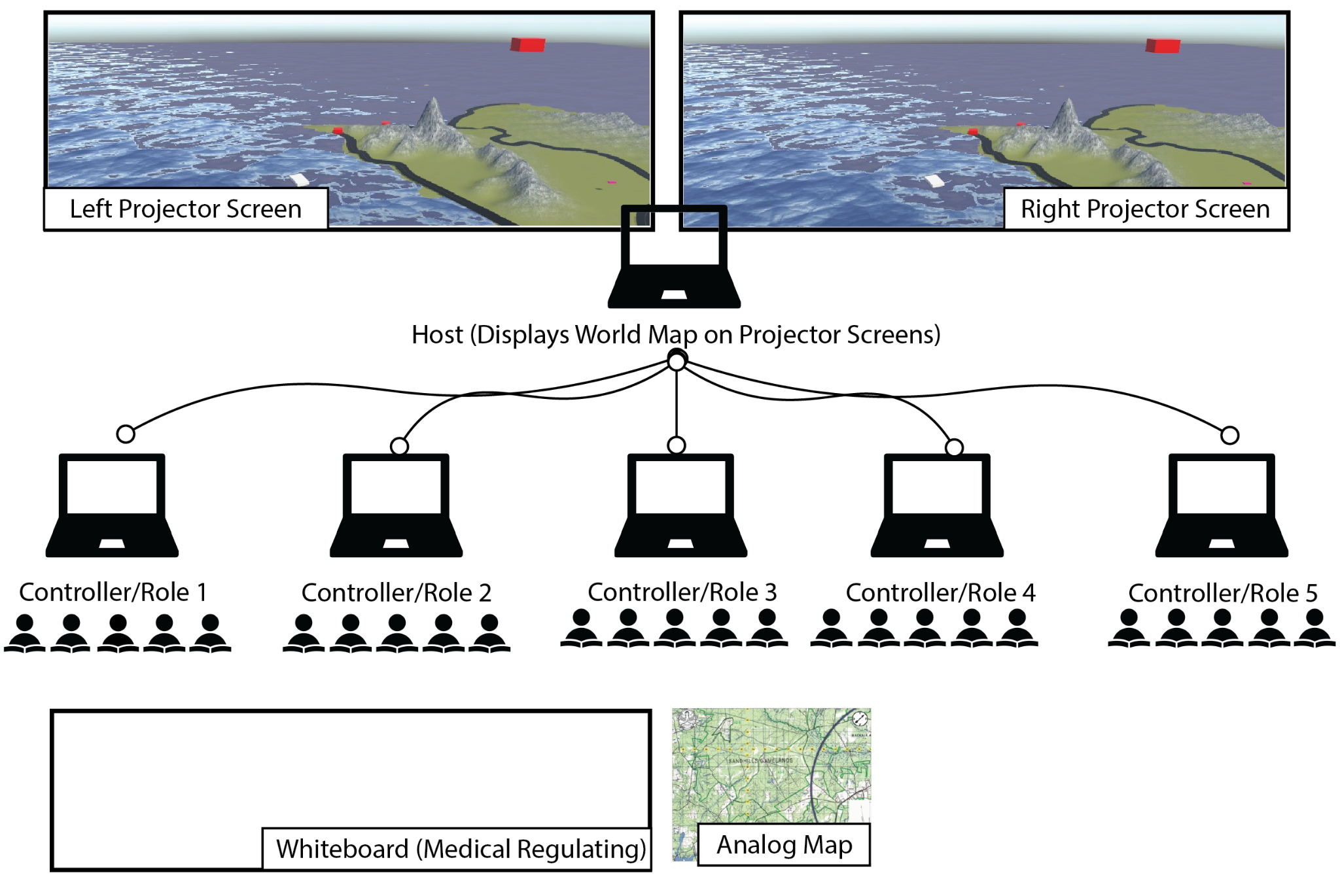}
\caption{The Medical Evacuation Doctrine Course classroom environment during execution.}
\label{fig:setting}
\end{figure}

Figure \ref{fig:setting} depicts the instructional setting, which may have between eight and thirty-three participants. Participants are provided with a large analog canvas map with magnetic unit icons, reflecting the same grid and terrain in the digital simulation. Participants are further provided with whiteboards to assist with medical regulating and may use digital devices to establish chat rooms and use platforms such as Excel, Google Sheets, etc. Seven to twelve participant laptops are loaded with the simulations, and each laptop is provided to a subgroup of participants assigned a position in the medical evacuation system. Each subgroup of participants assigns roles internally, including an individual that will move assigned platforms in the digital simulation in response to casualties and facilitator injects. All subgroups must collaborate to more optimally allocate resources and make evacuation decisions across an expansive battlefield. A command-and-control laptop displays the simulation world environment on projector screens and is primarily intended for facilitators and the simulation leadership. A dedicated facilitator laptop allows facilitators to add or remove casualties at will and facilitate casualty evacuation support. All laptops are wirelessly networked.

\section{Simulation Progression}

\begin{figure}[t]
    \centering
    \subfloat[\centering ]{{\includegraphics[width=0.49 \linewidth]{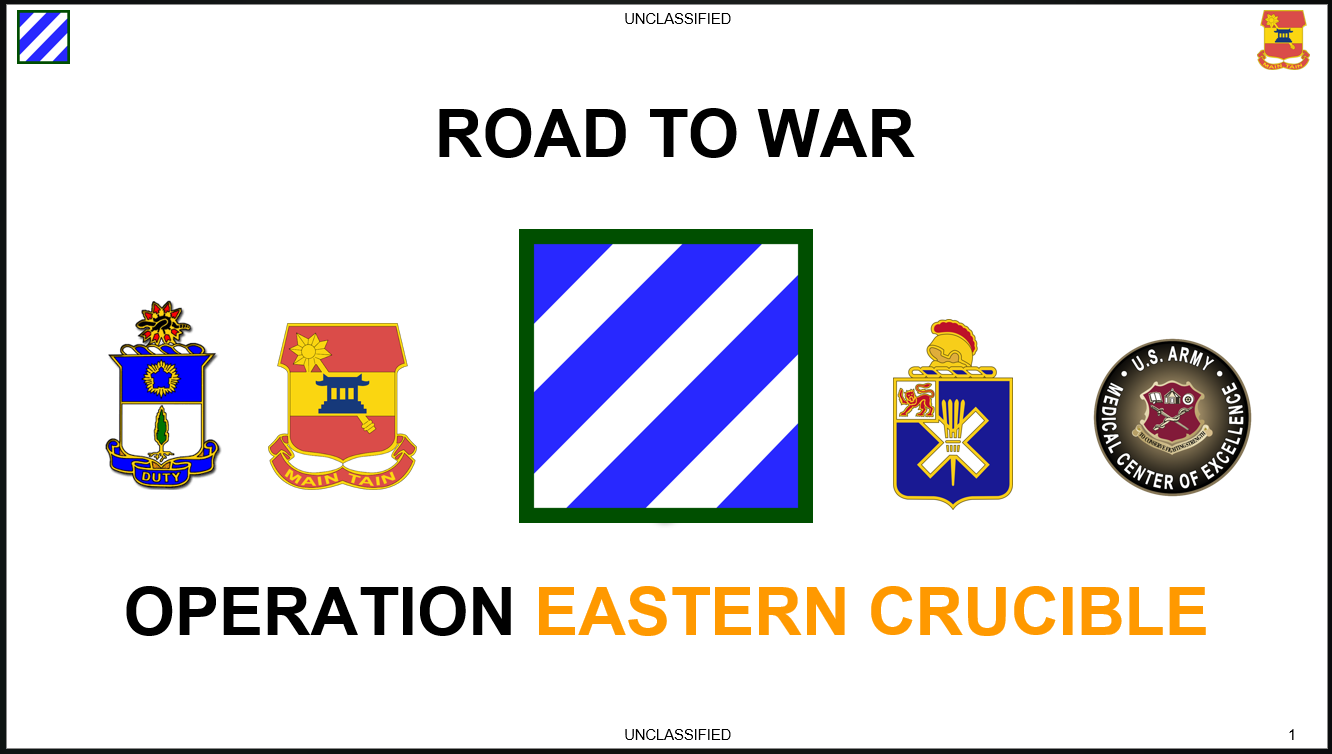} }}%
    \subfloat[\centering ]{{\includegraphics[width=0.49 \linewidth]{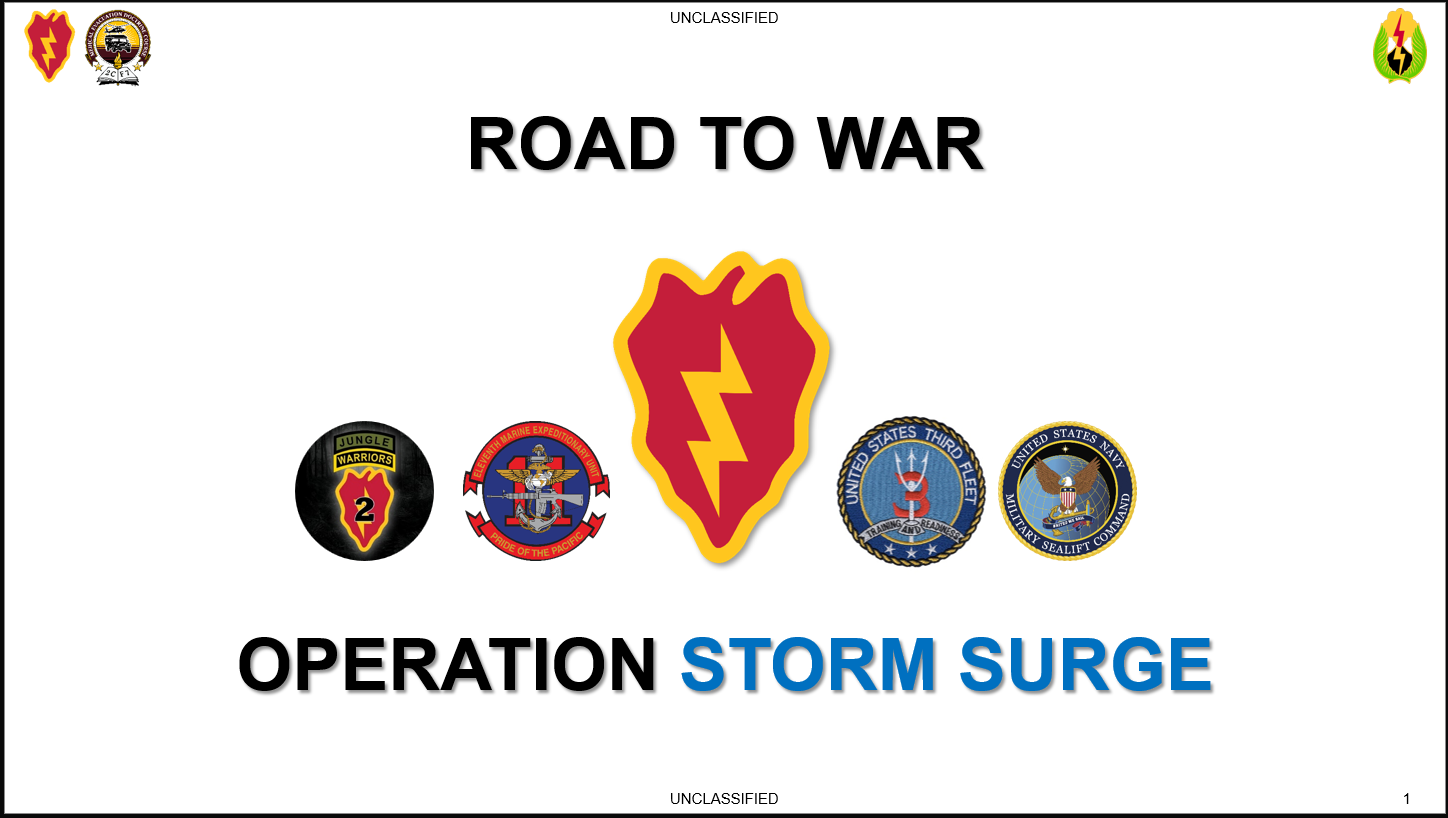} }}%
    \caption{Title slides for the road to war briefs provided to participants in the planning phase for both scenarios.}%
    \label{fig:OPORD}
\end{figure}

\begin{figure}[t]
    \centering
    \subfloat[\centering ]{{\includegraphics[width=0.495 \linewidth]{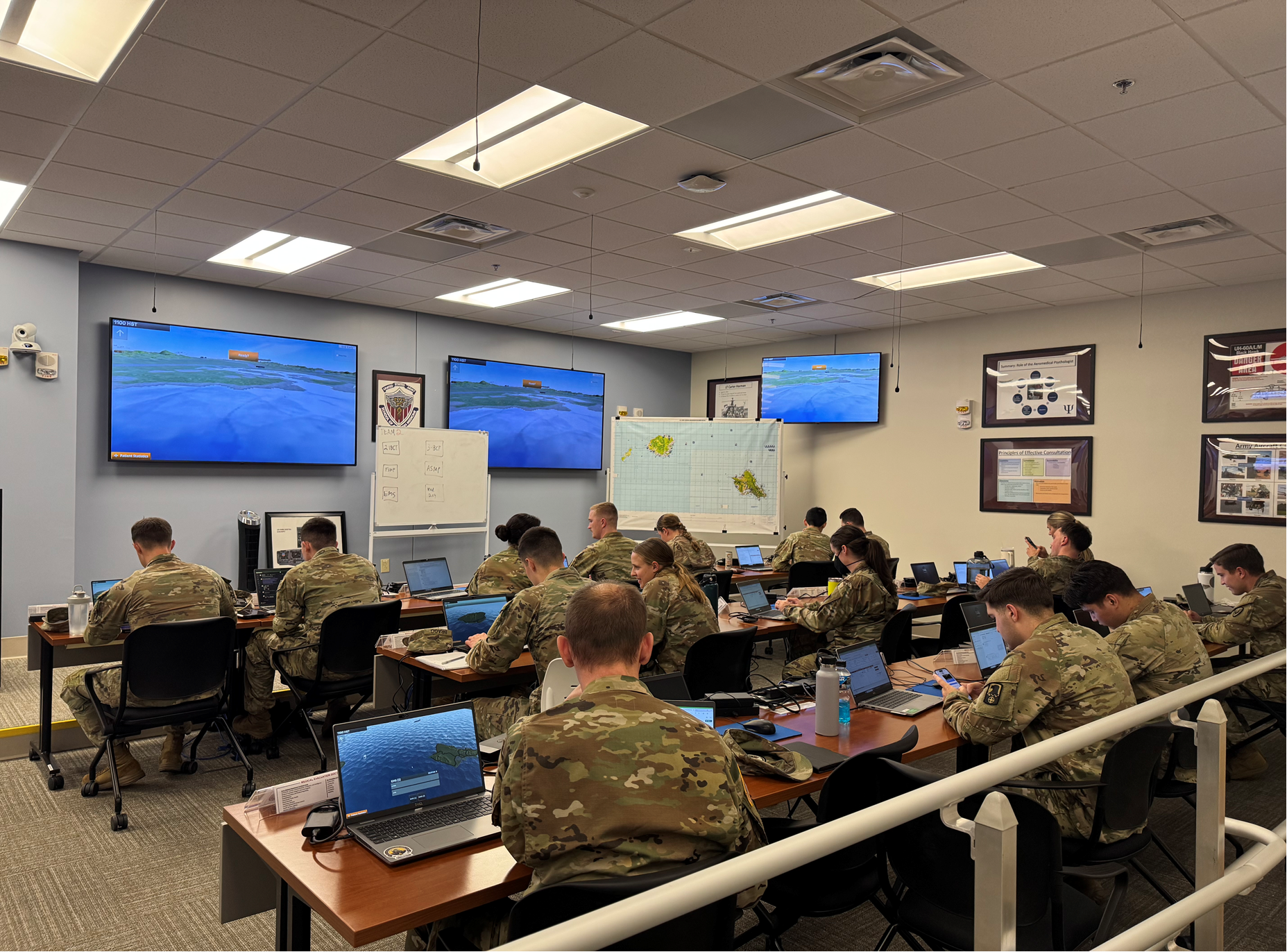} }}%
    \subfloat[\centering ]{{\includegraphics[width=0.49 \linewidth]{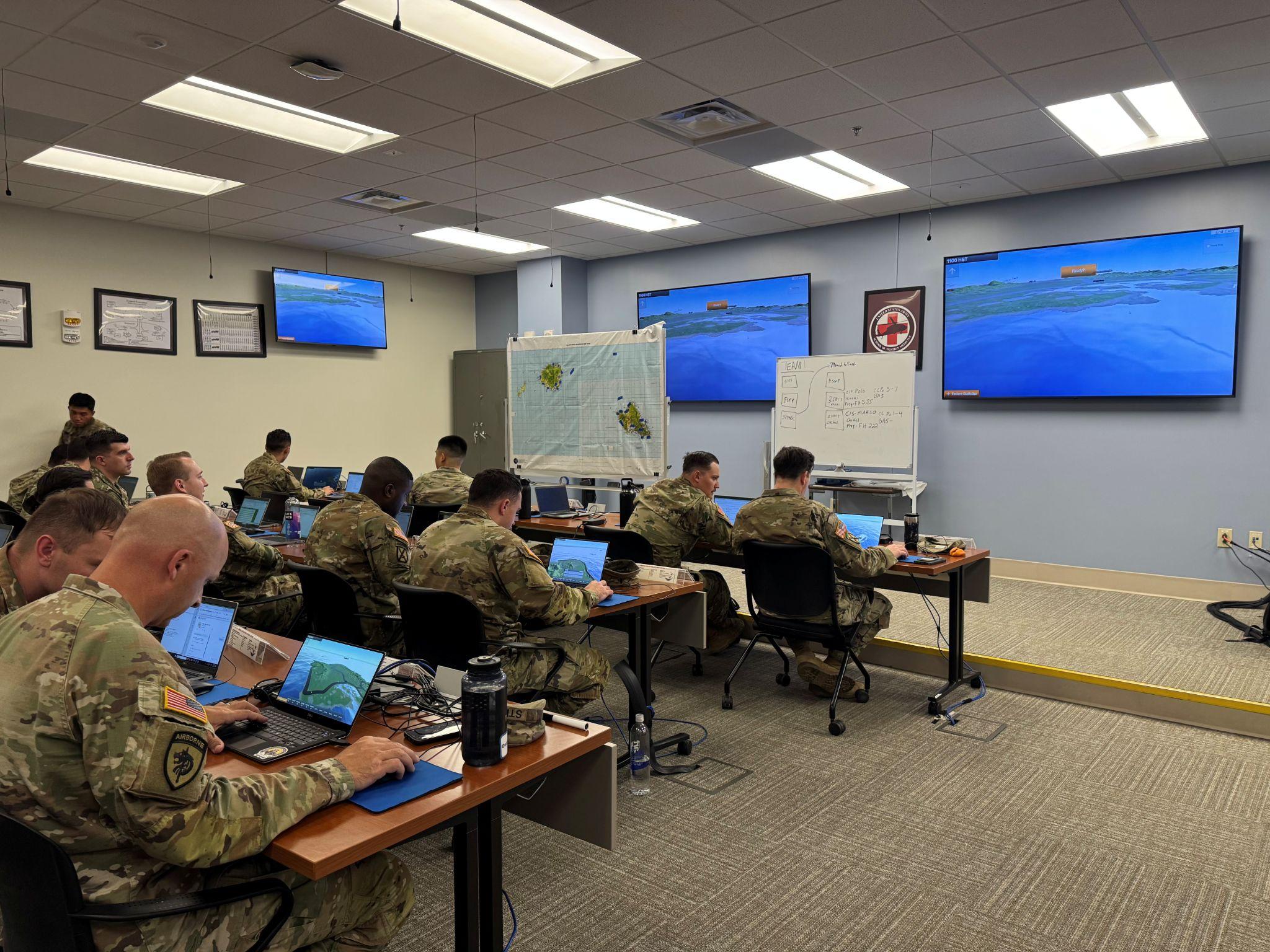} }}%
    \caption{Medical Evacuation Doctrine Course students participating in Operation Storm Surge.}%
    \label{fig:photos}
\end{figure}

Each simulation consists of three phases: planning, execution, and debriefing. Participants are assigned roles at the start of planning. Facilitators then provide participants with mission products, including an operations order and a road to war slide deck briefing, depicted in Figure \ref{fig:OPORD}. These products contain information on the mission timeline, casualty estimates, terrain description, and enemy threat considerations. In the planning phase, participants collaborate as one large team to determine the initial placement of AXPs, loading points, relay points, control points, and medical treatment facilities in accordance with the scenario. Participants have one to two hours to complete their planning and then conduct a hasty sustainment rehearsal, during which they brief the entire group on their platform allocation and command and control decisions by mission phase. Between planning and execution, facilitators update the location of each planned platform and facility in the digital simulation. The execution phase begins with a facilitator-driven simulation tutorial displayed, including how to dispatch evacuation platforms, transfer patients, and basic rotating, panning, and general movement control. The facilitators then officially start the simulation, and participants begin developing evacuation requests, regulating patients, and dispatching platforms in response to generated casualties and miscellaneous facilitator injects. The simulation may last for up to three hours, with each real-world hour reflecting six to eight in-game hours. Participants, therefore, play through up to twenty-four in-game hours in a single iteration. Figure \ref{fig:photos} shows Medical Evacuation Doctrine Course students in the planning and execution phases. After the simulation, the facilitators executed a data-driven half-hour group-wide debriefing, in which each phase of the simulation is discussed to identify lessons learned and issues experienced. Plots and graphs from the simulation are presented to participants to facilitate discussion. In the formal study, a fifteen-question questionnaire using a five-point Likert scale is provided to participants to assess participant learning outcomes and overall simulation quality. Similarly, an external observer is interviewed to help gauge participant reception to and performance in the simulation and identify friction points.

\section{Methodology}

\subsection{Research Design, Data Collection, and Ethics}

We quantitatively and qualitatively analyze two runs of the MEWI Operation Storm Surge scenario in a classroom setting to determine simulation effectiveness across a set of medical evacuation tasks and considerations. Specifically, we were interested in (1) the extent to which MEWI realistically represented the challenges of medical evacuation and regulating in a large-scale conflict, (2) the general playability and quality of the simulation, and (3) whether the simulation enabled key learning objectives related to medical regulating, triage, joint service integration, emerging evacuation platforms, and air-ground platform integration. The research design involved a classroom of 32 participants enrolled in the April 2025 254 Medical Evacuation Doctrine Course. The 32 participants were split into two competing teams, Team 1 and Team 2, and assigned roles. Most participants were either medical helicopter pilots or medical operations officers. At the time of their participation in the simulation, participants had received seven days of didactic training and medical evacuation planning practice. An external observer, who previously graduated from the Medical Evacuation Doctrine Course, was invited to take notes on participant participation in the simulation. Consenting participants were provided with paper survey forms to be filled out after the simulation. The Pacific simulation ran for ninety minutes. Data collected from the simulation included patient arrival times at treatment facilities, evacuation platform routing, location of patient deaths, and delay times stratified by patient precedence and type. All participants provided their written consent for the use of their survey responses for data analysis and in subsequent publications emerging from the simulation. Informed consent was collected in accordance with protocol title MEDWARGAME1 reviewed and approved for IRB-exemption (number 24-00039e) by the United States Army Medical Center of Excellence. All methods were carried out in accordance with relevant guidelines and regulations.

\section{Results}

\subsection{Simulation Data}

The following figures are associated with Team 1 but serve as a representative sample of participation results. We discuss Team 2 in instances where planning decisions diverged between teams.

\begin{figure}[t]
    \centering
    \subfloat[\centering ]{{\includegraphics[width=0.49 \linewidth]{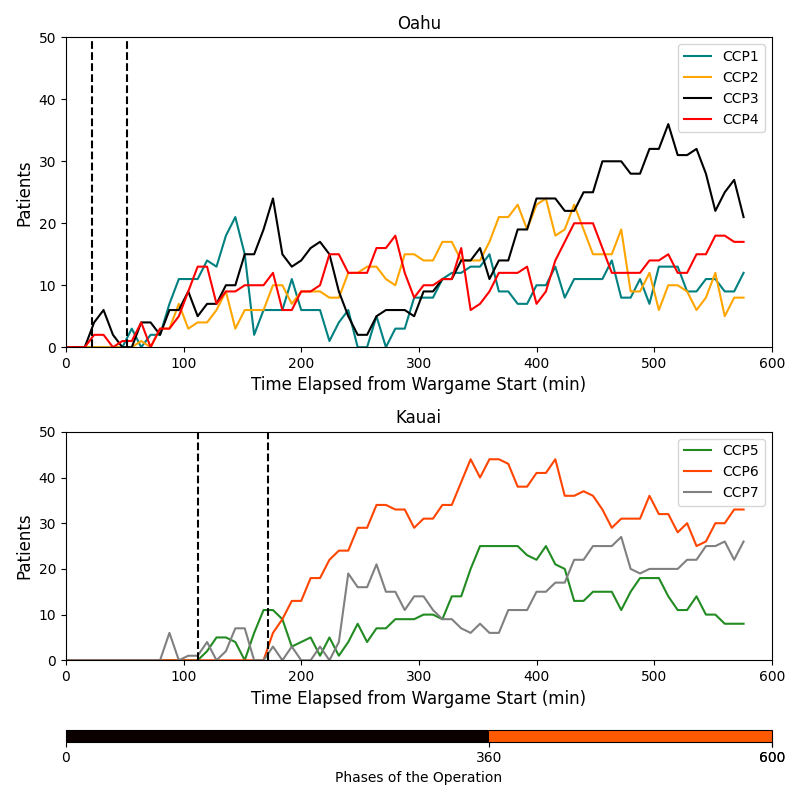} }}%
    \subfloat[\centering ]{{\includegraphics[width=0.49 \linewidth]{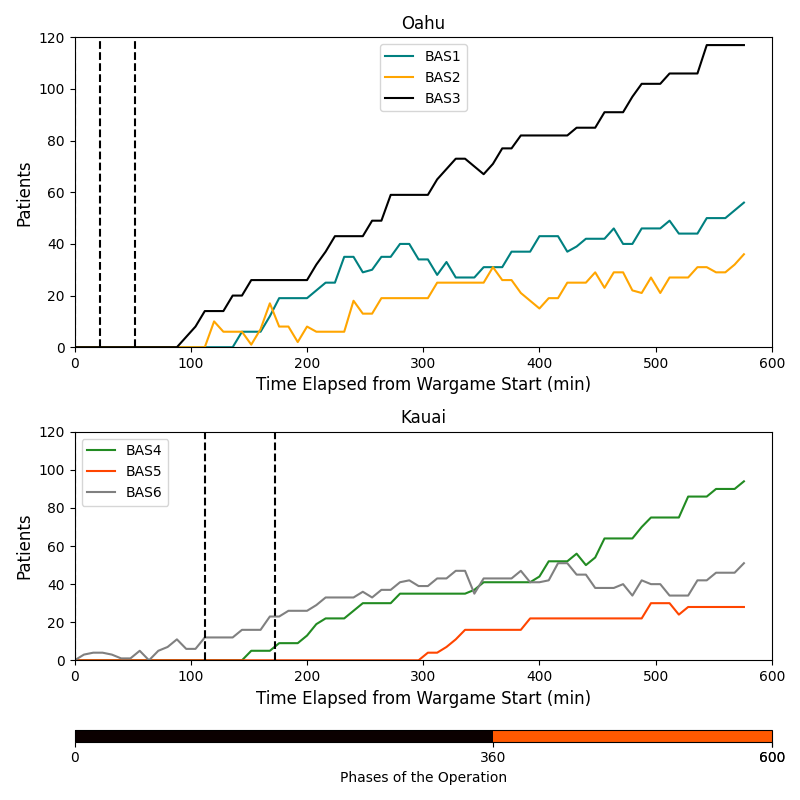} }}%
    \caption{Patient count at all casualty collection points (a) and Role 1 treatment facilities (b), also known as battalion aid stations, by island over time, for Operation Storm Surge.}%
    \label{fig:patientcount}
\end{figure}

Figure \ref{fig:patientcount} charts the total casualties at each CCP and Role 1 facility, also known as a battalion aid station (BAS), as the simulation progressed. The graphs reflect participants' ability to clear casualties from key nodes in the evacuation network. CCPs are sequentially activated at times represented by the dashed vertical lines. Throughout the simulation, casualties continuously accrue at CCPs at a rate informed by Poisson distributions. CCP clearance on Oahu remained uniform through the first 400 (in-game) minutes of the simulation, with an average of 12-15 patients waiting at each CCP at a given time. CCP 3, located on the easternmost side of Oahu, began to backlog significantly at 400 minutes. Due to communication issues relaying the evacuation request to the appropriate approval authorities, it would not be until 100 (in-game) minutes later that air support arrived to begin substantially clearing CCP3. Kauai experienced a more asymmetric clearance of on-island CCPs, with CCP 6, the northernmost CCP on Kauai, experiencing a mass casualty event almost immediately after activation. Despite the growing backlog at CCP 6, the on-island command and control authorities did not meaningfully employ air evacuation support to assist in clearance, and further did not reallocate additional ground evacuation platforms to the north. This teaching point was highlighted during the data-driven debriefing. Interestingly, air evacuation platforms were primarily used to clear Role 1 facilities on Kauai but primarily used to clear CCPs on Oahu. This reflects in the Role 1 clearance charts; casualties at BAS 3 (associated with CCP 3) on Oahu ballooned up to 120 patients at the end of the simulation, whereas casualties at BAS 6 (associated with CCP 6) rarely exceeded 20 patients at a time. More generally, CCP clearance was the primary focus of Team 1, with no greater than 45 patients waiting at any CCP at a given time. By contrast, the patient count increased nearly continuously at all Role 1 facilities.

\begin{figure}[t]
\centering
\includegraphics[width=0.60\linewidth]{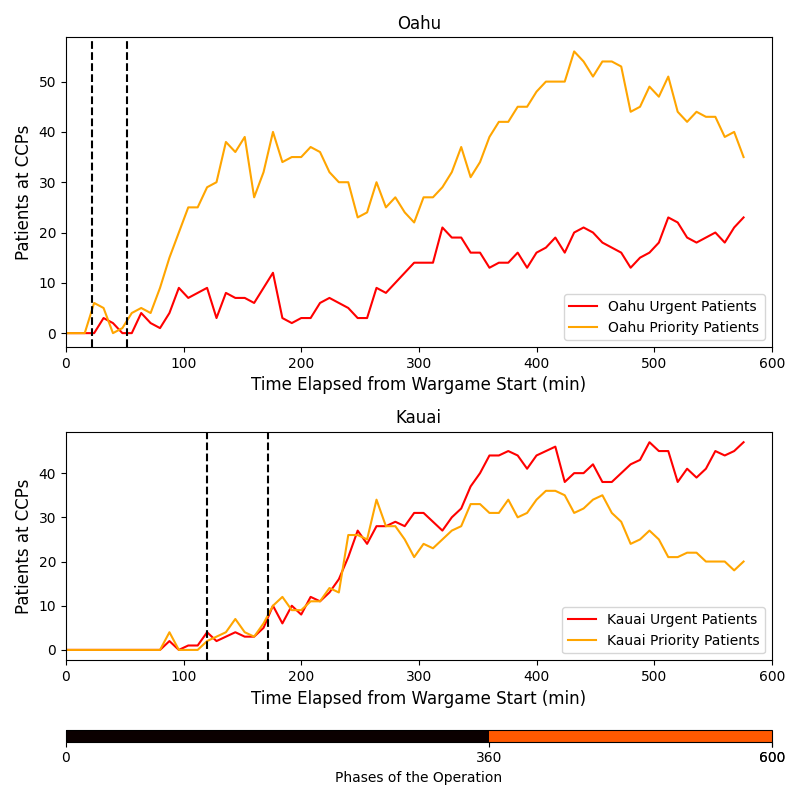}
\caption{Patient count stratified by precedence and summed across casualty collection points on each island over time, for Operation Storm Surge.}
\label{fig:stratified}
\end{figure}

Figure \ref{fig:stratified} highlights patient precedence in accumulating patient counts across all CCPs as the simulation progressed and reflects how well simulation participants triaged patients before transport. Urgent patients, shown in red, have a substantially higher mortality rate than priority patients, shown in orange, and must be prioritized for evacuation. Simulation participants on Oahu appropriately sequenced urgent patients ahead of priority patients for evacuation, whereas simulation participants on Kauai did not distinguish by patient precedence, which served as another talking point for the data-driven debriefing.

\begin{figure}[t]
\centering
\includegraphics[width=0.60\linewidth]{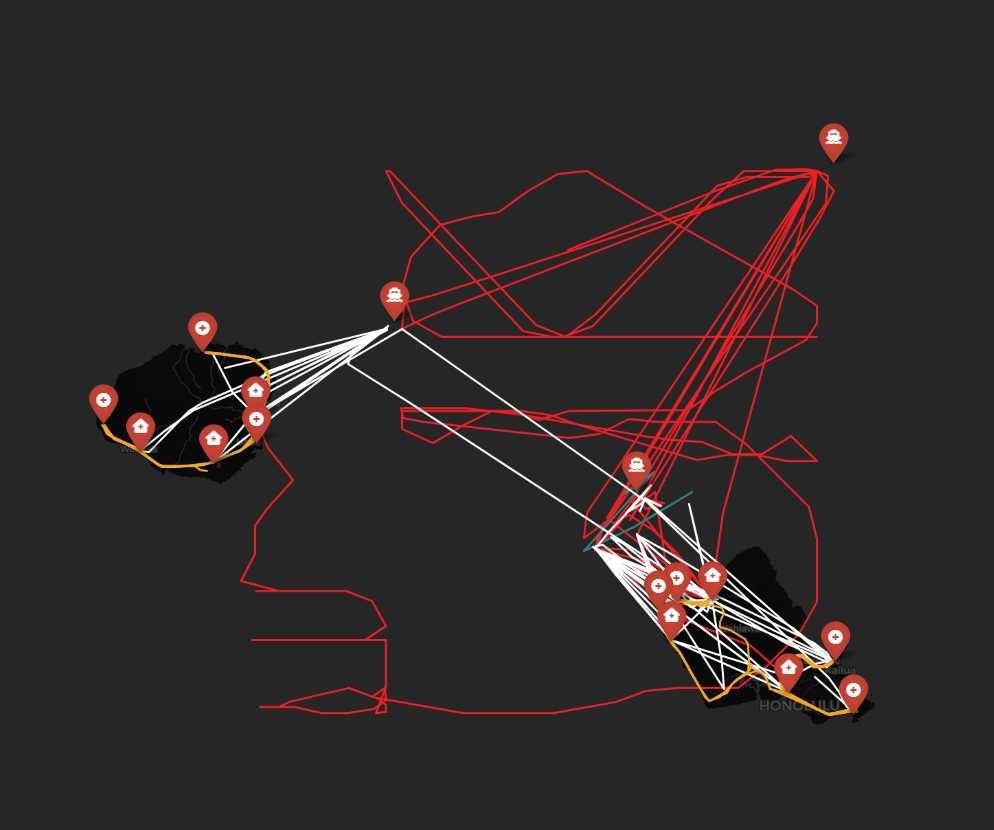}
\caption{Medical evacuation platform routing over time across islands, for Operation Storm Surge.}
\label{fig:overview}
\end{figure}

\begin{figure}[t]
    \centering
    \subfloat[\centering ]{{\includegraphics[width=0.505 \linewidth]{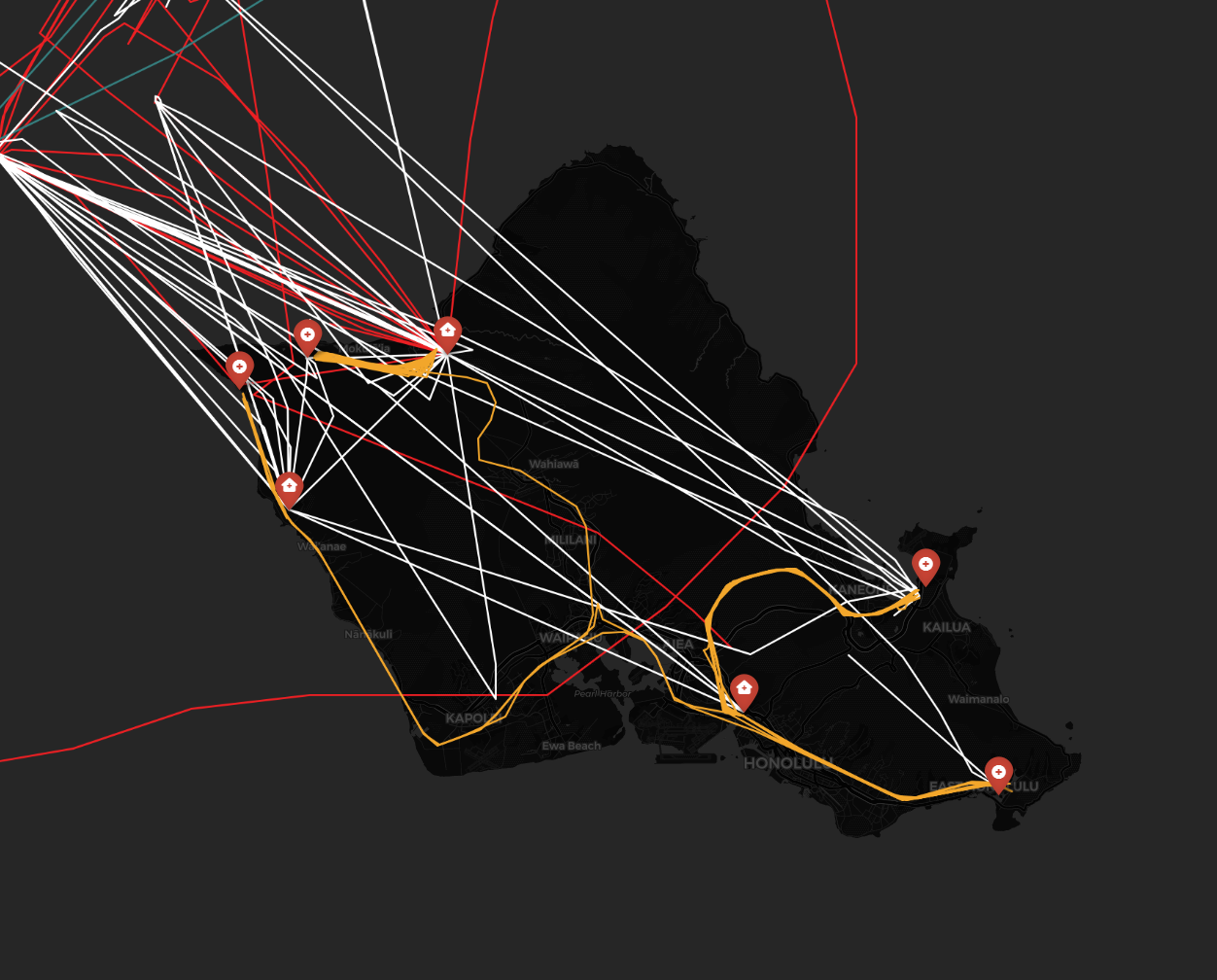} }}%
    \subfloat[\centering ]{{\includegraphics[width=0.485 \linewidth]{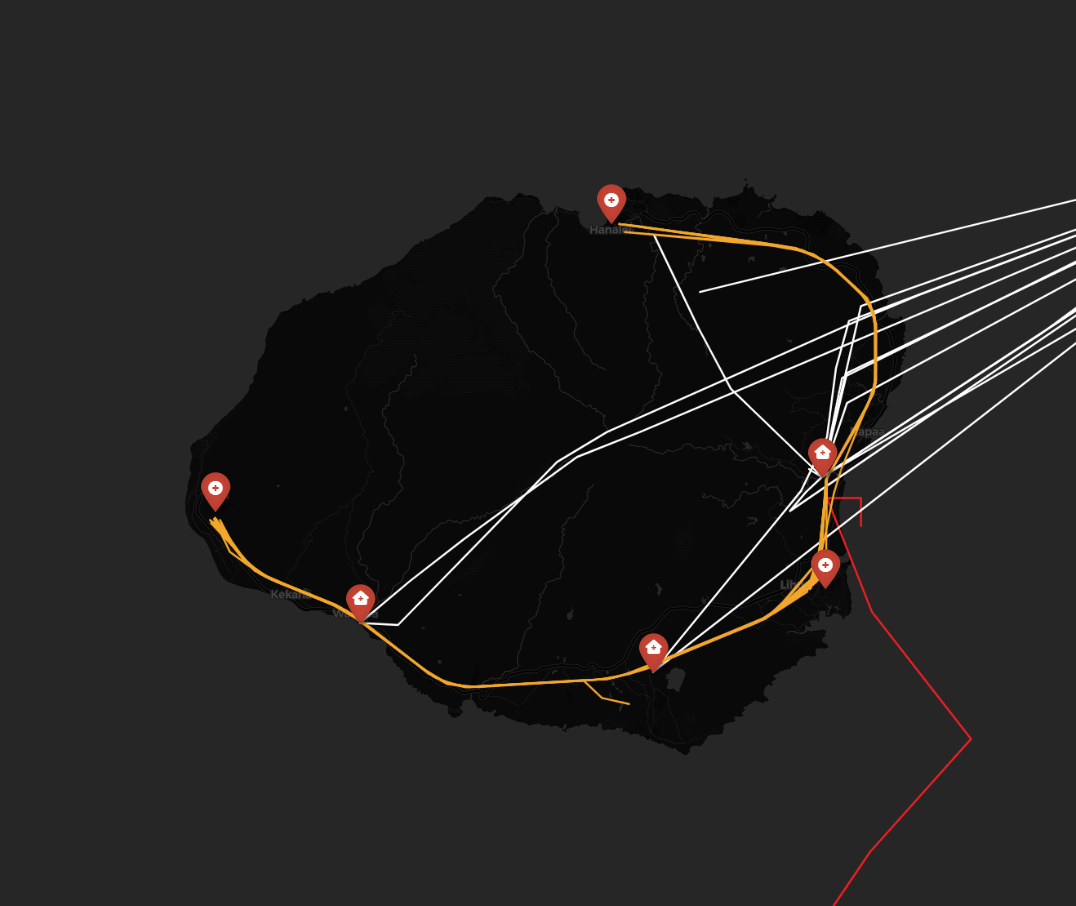} }}%
    \caption{Magnified view of medical evacuation platform routing over time for the islands of Oahu (a) and Kauai (b), for Operation Storm Surge.}%
    \label{fig:magnified}
\end{figure}

Medical evacuation platform routing is shown in Figures \ref{fig:overview} and \ref{fig:magnified}. The Mercy hospital ship, the scenario Role 3 facility, is fixed in place, located in the top right of Figure \ref{fig:overview}, and is marked with a red ship icon. Also marked with red ship icons are the two EMS, serving as the scenario's Role 2s. Team 1 simulation participants initialized the first EMS north-east of Kauai and the second EMS to the north-west of Oahu. The Kauai EMS remained in place throughout the simulation, whereas the Oahu EMS traversed a route parallel to the northern coast of the island, indicated in blue. The routing structure for the ASMP with MV-75 helicopters is shown in red and indicates that ASMP aircraft were primarily used to move patients from Role 2 EMS to the Role 3 hospital ship. The routing structure for the FSMP with HH-60M helicopters is shown in white and indicates that FSMP aircraft were primarily used to move patients from the on-island Role 1s to the Role 2 EMS. Two FSMP aircraft were allocated to Oahu, while one FSMP aircraft was allocated to Kauai. At no point were FSMP aircraft reallocated between islands. Notably, the Role 2 EMS were placed near the islands, shifting the brunt of the air evacuation load to the ASMP rather than the FSMP. Red icons with houses reflect Role 1 facilities, and red icons with circles reflect CCPs. Ground evacuation platform routing is shown in orange. Figure \ref{fig:magnified} presents the same position data as Figure \ref{fig:overview}, but separated between islands, to more clearly showcase evacuation platform employment.

\begin{figure}[t]
\centering
\includegraphics[width=0.70\linewidth]{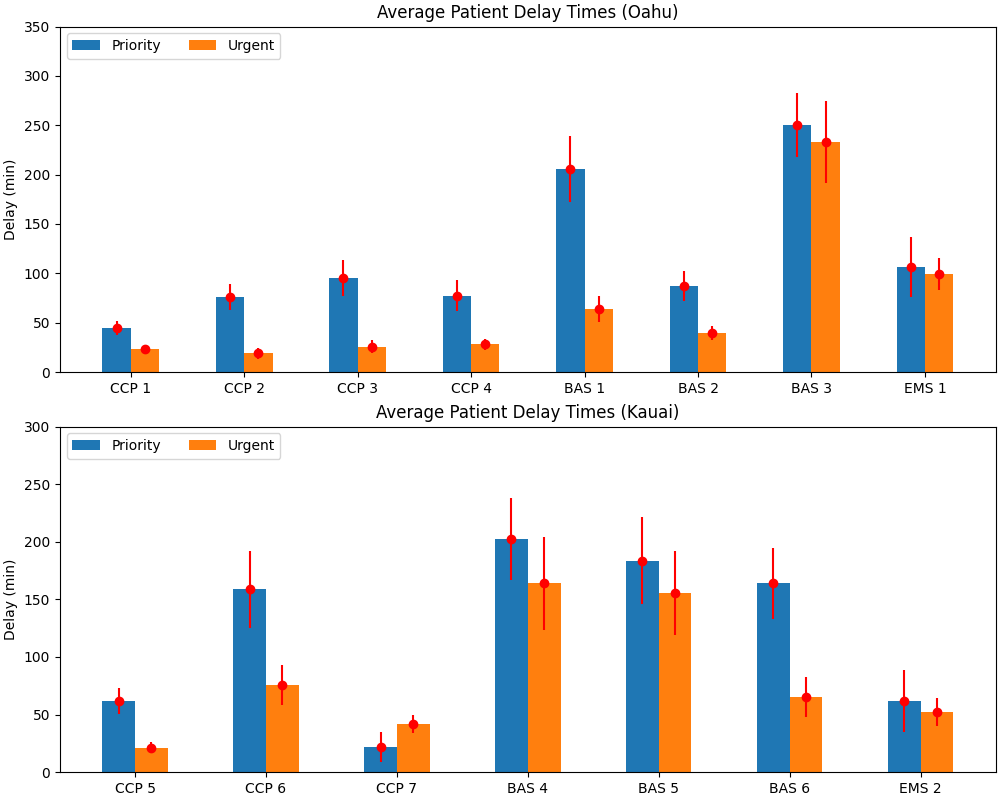}
\caption{Average patient delay by precedence by island, for each casualty collection point or role of care, with 95\% confidence intervals, for Operation Storm Surge.}
\label{fig:precedence}
\end{figure}

Figure \ref{fig:precedence} presents average patient delay by precedence, stratified by island and across CCPs and roles of care. A delay is defined as the time a patient spends at a medical facility after having received treatment but before being picked up by an evacuation platform, or for CCPs, the time a patient spends after initializing but prior to initial pick up by an evacuation platform. Much like Figure \ref{fig:stratified}, Figure \ref{fig:precedence} signals triage performance, now separated by individual nodes in the evacuation network. Delay times for urgent patients at all Oahu CCPs and BAS 1 and BAS 2 were significantly less than delay times for priority patients; this indicates that simulation participants on Oahu performed adequate triage at those locations. Oahu simulation participants failed to meaningfully triage at BAS 3 or from the Oahu-affiliated Role 2 EMS. Similarly, simulation participants on Kauai performed adequate triage at CCP 5, CCP 6, and BAS 6, but failed to do so at CCP 7, BAS 4, BAS 5, or the Kauai-affiliated Role 2 EMS.

\begin{figure}[t]
\centering
\includegraphics[width=0.70\linewidth]{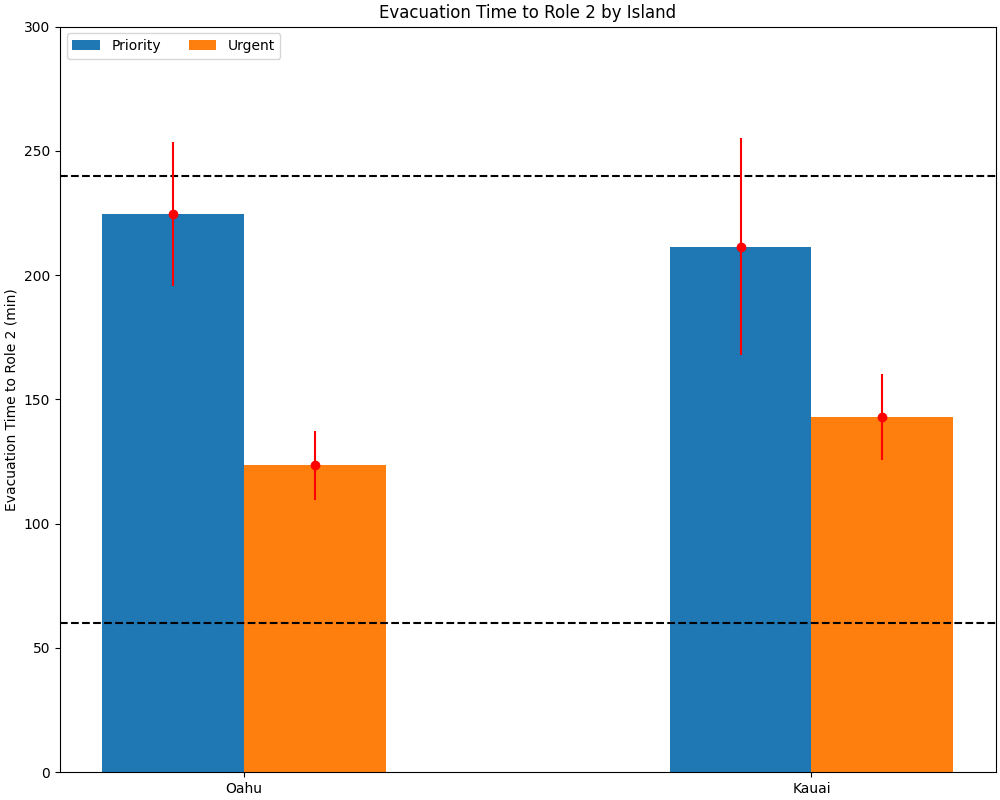}
\caption{Evacuation times to Role Two treatment facilities, by patient precedence, with 95\% confidence intervals, by island, for Operation Storm Surge.}
\label{fig:evactimes}
\end{figure}

Figure \ref{fig:evactimes} shows the cumulative time a patient takes to arrive at a Role 2 for treatment, by island and by precedence. The horizontal dashed lines at 60 and 240 minutes reflect the evacuation standard for urgent and priority patients, respectively. The average evacuation times for urgent patients on both islands far exceeded the urgent patient evacuation standard, whereas the average evacuation times for priority patients on both islands met the priority patient evacuation standard. On both islands, urgent patients arrived at the Role 2 significantly faster than priority patients, indicating that the participants effectively employed triage. There was no significant difference in evacuation times for priority patients between islands, and no significant difference in evacuation times for urgent patients between islands. This generally suggests that the distribution of evacuation platforms on Oahu and Kauai was appropriately balanced against the future demand for patient movement.

\begin{figure}[t]
\centering
\includegraphics[width=0.70\linewidth]{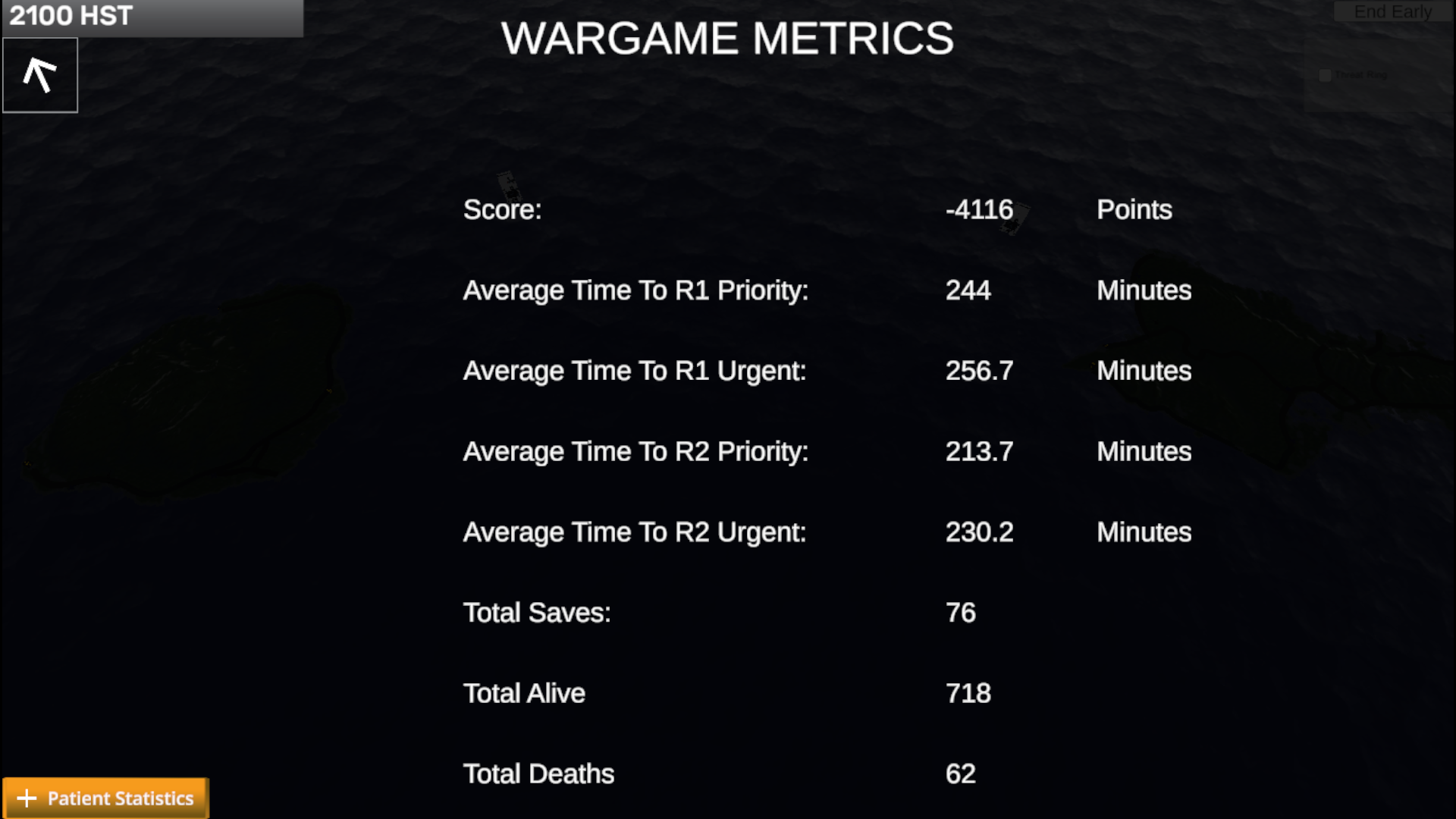}
\caption{Example of the post-simulation quick snapshot scoring screen.}
\label{fig:metrics}
\end{figure}

Figure \ref{fig:metrics} depicts a scoring screen shown to simulation participants and facilitators post-simulation, which helps facilitate later discussion. The score itself serves as a high-level metric for facilitators to gauge relative performance between classes, as participants are likely to participate in each simulation scenario only once. Total saves indicate the number of patients who arrive at the Role 3 hospital ship, and total alive indicates the number of patients who are still transiting the evacuation network.

\subsection{External Observer Comments} External observation focused on performance across the two simultaneously operating teams. Key scenario injects included a communications blackout from 1300--1330 and a 15-patient mass casualty (MASCAL) event at Kauai CCP 7 at 1500. Team 1, shown in the results section, encountered challenges in air movement coordination, resulting in poor patient transport and bottlenecks at EMS locations. Despite keeping untreated casualty counts low early, the signs of MASCAL conditions went unrecognized, and prior planning only accounted for single-island surges. Participants accepted the risk of fully unregulated patient movement via air to relieve MASCAL conditions, dedicating Bell MV-75s forward to offload Role 1 facilities. Terrain limitations were later identified, with Oahu citing vast distances and Kauai identifying the southern-focused BAS placement as key constraints. Team 2 demonstrated more structured planning, adhering to the Rotary Wing Utilization Plan and maintaining more organized patient regulation. While recognizing the MASCAL, Team 2 did not initiate a MASCAL plan at first. Their eventual response to the MASCAL on Kauai improved local conditions but unintentionally caused overflow on Oahu. The main limiting factor on Kauai was identified as the limited ability of air assets to clear out Role 1 facilities. Team 2's approach had more structure and quicker responsiveness overall, while Team 1's reactive posture led to operational strain. Both teams struggled with MASCAL response, highlighting the need for more rigorous planning.

\subsection{Participant Survey Data}

\begin{figure}[t]
\centering
\includegraphics[width=0.99\linewidth]{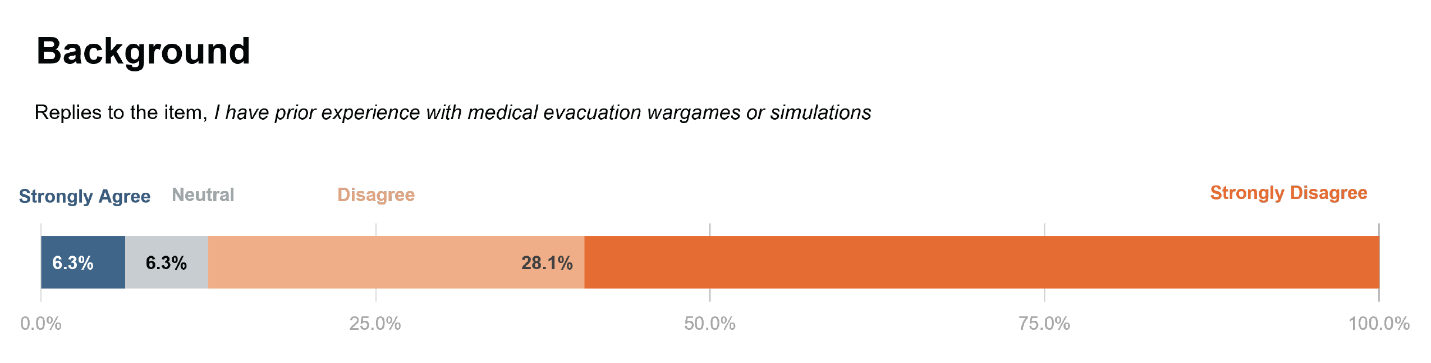}
\caption{Participant survey data on medical evacuation wargame or simulation experience.}
\label{fig:simexperience}
\end{figure}

\begin{figure}[t]
\centering
\includegraphics[width=0.99\linewidth]{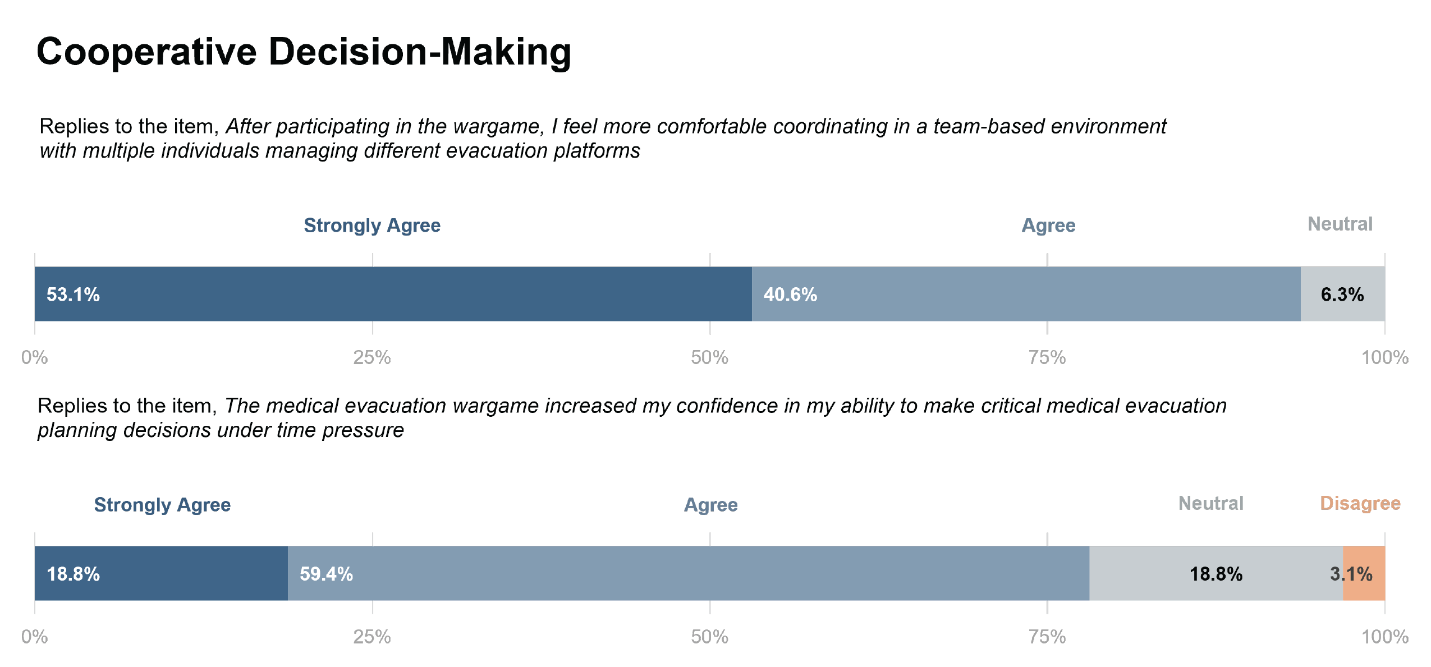}
\caption{Participant survey data on cooperative decision-making after intervention.}
\label{fig:decisionmaking}
\end{figure}

\begin{figure}[t]
\centering
\includegraphics[width=0.99\linewidth]{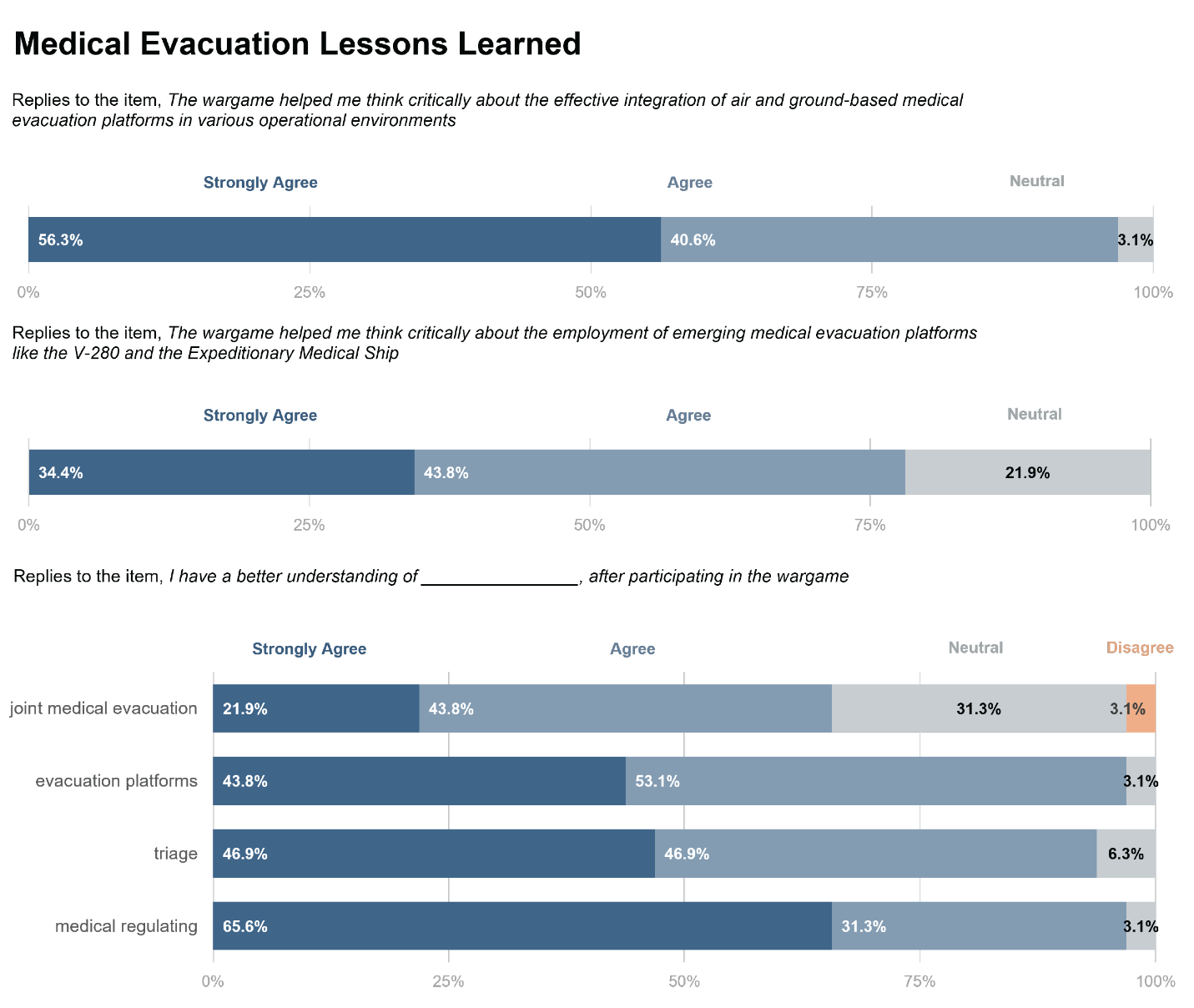}
\caption{Participant survey data on medical evacuation lessons learned after intervention.}
\label{fig:lessonslearned}
\end{figure}

\begin{figure}[t]
\centering
\includegraphics[width=0.99\linewidth]{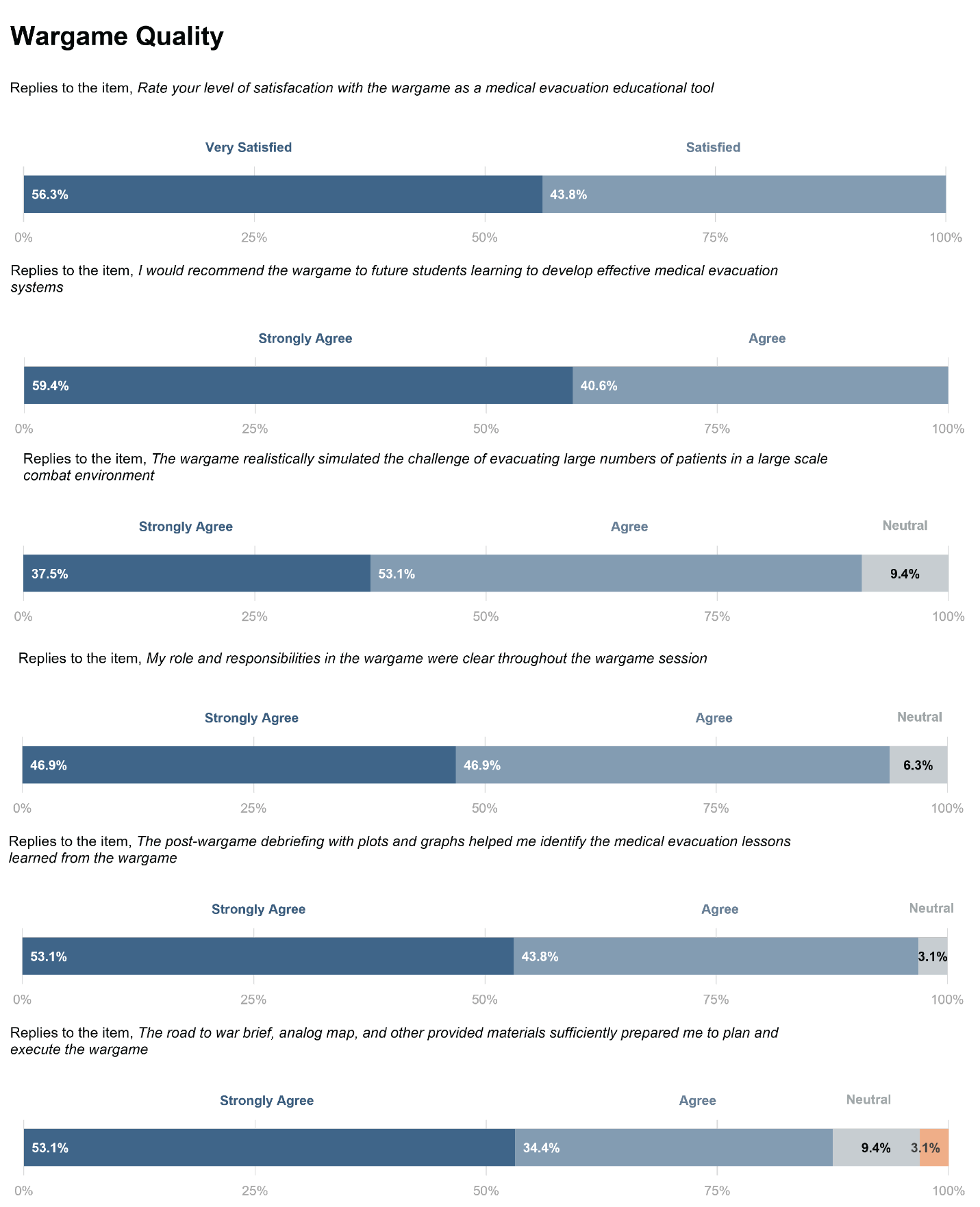}
\caption{Participant survey data on medical evacuation simulation wargame quality.}
\label{fig:wargamequality}
\end{figure}

The post-simulation Likert survey contained fifteen questions organized into four categories: background, cooperative decision-making, medical evacuation lessons learned, and simulation quality. As seen in Figure \ref{fig:simexperience}, simulation participants had very little experience with medical evacuation simulations or simulation of any kind prior to interacting with MEWI. Figure \ref{fig:decisionmaking} considers domain-agnostic decision-making and team coordination. Most participants (53.1\%) strongly agreed that participating in the simulation made them feel more comfortable coordinating medical evacuation efforts in a group setting, whereas most participants (78.2\%) either strongly agreed or agreed that participating in the simulation increased their confidence in their ability to make critical medical evacuation planning decisions under time pressure. Figure \ref{fig:lessonslearned} examines the simulation impact on six key medical-evacuation focus areas stressed during didactic training: air-ground medical evacuation integration, emerging medical evacuation platforms, joint medical evacuation, medical evacuation platform constraints and capabilities, triage, and medical regulating. A substantial 65.6\% of participants strongly agreed that the simulation helped them understand medical regulating -- considered by the Medical Evacuation Doctrine Course facilitators to be one of the most challenging tasks to perform, and a known weakness in operational units. More than 95\% of participants either agreed or strongly agreed that the simulation furthered learning related to medical evacuation platform constraints and capabilities, triage, and air-ground medical evacuation integration. Fewer participants felt that the simulation furthered learning related to joint medical evacuation or emerging evacuation platforms; 21.9\% and 31.3\% of participants marked neutral for those two categories, respectively. This may be due to limited touch-points with the material during didactic training, or due to the limited number of players in the Pacific scenario that directly interfaced with either the joint medical evacuation platforms (EMS and hospital ship) or the emerging evacuation platforms (EMS and Bell MV-75). Finally, Figure \ref{fig:wargamequality} depicts outcomes related to overall simulation quality. A majority of participants (56.3\%) reported being very satisfied with the simulation, and 59.3\% strongly agreed that they would recommend it to future participants learning to develop effective medical evacuation systems. Additionally, 37.5\% agreed and 53.1\% strongly agreed that the simulation realistically represented the challenge of evacuating large numbers of patients in a large-scale conflict. Participants spoke highly of the simulation structure and training materials, with special emphasis placed on the clarity of roles and responsibilities, the value of the data-driven debriefings, and the quality of provided materials to help facilitate simulation planning.

\section{Discussion}

The simulation iterations demonstrate key trends in real-world medical evacuation employment, particularly in casualty triage, air-ground integration, and platform routing. Figure \ref{fig:patientcount} reveals that urgent patients were consistently prioritized across both islands, while priority patients accumulated at CCPs and Role 1s over time. This pattern aligns with doctrinal expectations for triage as well as in-game incentives. Yet, as shown in Figure \ref{fig:evactimes}, on average, the movement of priority patients to the Role 2 still met the 240-minute priority evacuation standard while the movement of urgent patients significantly exceeded the 60-minute urgent evacuation standard. Six of seven CCPs significantly prioritized urgent above priority patients, while only three of six Role 1s significantly prioritized urgent above priority patients. This suggests more focus needs to be placed on evacuating urgent patients at the battalion aid stations, and that meeting the 60-minute urgent evacuation standard may be impossible for the given scenario and associated casualty numbers. The latter aligns with real-world discussion about the military's inability to meet the one-hour evacuation standard in future large-scale combat operations with current resources, manning, and adversarial threats.

The map's geography also influenced evacuation performance. As Figure \ref{fig:patientcount} illustrates, CCP 3 and CCP 6 showed steady growth in patient count, implying limited coordination and planning for these more remote locations. For reference, CCP 6 is the northernmost CCP in Kauai, and as shown in Figure \ref{fig:magnified}, it is the most disconnected and primarily uses ground transport, which is the least efficient. In contrast, CCPs 2 and 5 saw eventual declines, likely due to proper planning. CCP 2 is in the top left of Oahu, and Figure \ref{fig:magnified} shows how interconnected it is with aerial platforms. The spatial mismatch between asset placement and casualty load was more pronounced on Kauai, where Role 1 facilities were clustered in the south. Participants did not anticipate this until well into the scenario, delaying response adaptation. Of note was the MASCAL event, which caused a brief spike in CCP 7, but despite CCP 7 being settled, there is a general trend upward in-patient delay on the other CCP's likely due to reallocation of resources towards the MASCAL.

While aggregate platform utilization rates were not explicitly tabulated, Figures \ref{fig:overview} and \ref{fig:magnified} suggest a potential over-reliance on air platforms for casualty movement. Participants tended to prioritize shifting high-speed, high-capacity air platforms forward while overlooking slower ground platforms, evident from the light use of orange ground routing lines. The external observer noted the presence of up to 40\% of ground platforms across both islands being idle at any given time. This was despite 79\% of all deaths occurring prior to reaching the on-island Role 1. Similarly, rather than employ an ambulance shuttle system to transport patients closer to the Role 2 EMS, participants often over-extended air platforms to the far side of both islands. This trend matches a real-world friction point in military medical evacuation --- the historical over-reliance on air evacuation and lack of meaningful air-ground evacuation platform integration. It is worth mentioning that aerial platforms in combat scenarios are more constrained than the simulation presents. Fuel constraints, enemy uncertainty, adequate landing zones, and more make aerial evacuation more challenging, which leads to the need for a robust ground evacuation network. External observer comments mention aircraft idling and poor coordination of evacuation requests, which conflicted with the team's original rotary-wing utilization plan.

Overall, the participants performed well, given the intentionally overwhelming nature of the simulation. While these insights are drawn from two simulation iterations, they reflect broader trends in medical evacuation planning and operational execution. Continuity of care was enforced, and participants made evacuation platform placement decisions in response to the evolving and substantial flow of casualties. Figures \ref{fig:overview} and \ref{fig:magnified} show the rotary platforms' tendency to stay close to the shoreline facing the Role 3 hospital ship, as they know the most consistent path will be between an island, to the Role 2 EMS, and then to the Role 3. This evacuation process relies heavily on air power, as participants were hesitant to move the EMS platforms, shown by the minimal routing structure in blue. This cycle helped maintain the continuity of care but also explains why CCPs on the outskirts of the island accumulated more casualties, and almost only relied on ground vehicles.

\subsection{Limitations \& Future Work} One limitation of this research study is the need for repeatability to assess learning retention and improvement. Due to the intensive and fast-paced nature of military schoolhouses, participants participated in each simulation scenario only once. While the Likert surveys measured learning, participants did not have the opportunity to replay the same scenarios and apply lessons learned in quick succession. This limits the assessment of skill progression within the simulation. While quantitative scoring can track improvements in the medical evacuation application, the absence of repeated play introduces ambiguity in evaluating the game's effectiveness as a learning tool. However, this does not diminish the simulation's instructional value. Army training is an ongoing process, and participants will apply similar decision-making skills in future unit training and operational planning. Future studies will address this limitation by subsequently evaluating participants in field settings and monitoring whether the simulation leads to measurable improvements in tactical performance. 

Another limitation was the small sample size. The empirical results presented are derived from a pilot study involving a single cohort of 32 students. As such, the quantitative metrics regarding evacuation efficiency should be viewed as illustrative of the simulation's capability to capture trends, rather than statistically generalized conclusions. For example, the casualty accumulation and triage documented in Figures \ref{fig:patientcount} and \ref{fig:stratified} are sensitive to scenario-specific casualty generation rates and CCP activation patterns. Similarly, routing and evacuation times in Figures \ref{fig:overview}-\ref{fig:evactimes} reflect outcomes driven by geography and evacuation platforms unique to each scenario, limiting direct cross-scenario comparison. Future work will involve multiple iterations of the same scenario across different student cohorts to isolate the effects of specific environment characteristics.

The simulation lacks a few features, currently in development, that meaningfully affect medical evacuation decision-making. Fuel and fighter management considerations limit the employment of evacuation platforms to specific ranges and time windows and can induce substantial delays if not appropriately planned. The adversarial intelligent agent, while functional, remains in development. An improved intelligent enemy will provide valuable insights into evacuation planning around threats, increase simulation difficulty, emphasize the need for participant communication, and add realism. Although medical evacuation platforms are protected under the Geneva Convention and Law of Armed Conflict, medical evacuation operators are trained to  expect adversaries who may not respect these protections (Geneva Convention I Arts. 35--36, 1949).\cite{yingling1952geneva} Enhancing casualty generation using real-world data and self-exciting stochastic processes may further improve realism.

A final limitation of the study is the absence of automated baseline policies (e.g., a heuristic, myopic, or optimal solver) against which participant performance can be quantitatively benchmarked. MEWI is structured for human-in-the-loop decision-making rather than algorithm development. To address this, we are developing the \textit{MEDEVAC-X} testbed, a MEWI extension for implementing and evaluating semi-Markov multi-agent decision-making algorithms. Future studies will use MEDEVAC-X to establish reference policies, enabling rigorous comparison between human strategic planning and algorithmic optima.

\section{Conclusion}

Wargaming has been an essential component of military preparation for centuries. In partnership with the United States Army, our study extends, for the first time, digital simulations to military medical evacuation to enable robust decision-making under uncertainty and reinforce doctrinal medical evacuation tactics and techniques. We coordinated with the Medical Evacuation Doctrine Course, the Army's only course dedicated entirely to medical evacuation and regulation, to exercise multiple iterations of a game-engine based medical evacuation simulation with overwhelming numbers of casualties. The simulation stressed participant planning, preparedness, and coordination for realistic operational scenarios in Europe and the Pacific. Multi-modal evacuation platforms with realistic patient-carrying constraints, mortality curves, and casualty generation reinforced the realism of the game and permitted facilitators to assess participant practical knowledge across multiple learning objectives. Experimental simulation data showed that participants adhered to key doctrinal principles and fell into common real-world traps, like an over-reliance on air power, which facilitators discussed in post-simulation data-driven debriefings. Participant deficiencies in the simulation helped course facilitators identify gaps in didactic training. Finally, post-simulation surveys indicated that the simulation was an effective tool for thinking critically about air-ground evacuation platform integration, casualty and medical evacuation integration, triage, and medical regulating, and that the simulation itself was realistic and of high-quality.

\bibliographystyle{SageV}
\bibliography{references}

\end{document}